%% file: _main.tex
\input{_constants}
\cameraready % \review OR \arxiv OR \cameraready
\pdfoutput=1
\documentclass[10pt,twocolumn,letterpaper]{article}
\input{cvpr_header}

\unless\ifarxiv \myexternaldocument{_supplementary} \fi

\usepackage{footmisc}

\begin{document}

%% TITLE
\title{\paperTitle}
\author{\authorBlock}
\maketitle

\renewcommand{\thefootnote}{*}
\footnotetext[1]{Corresponding author.} 

\input{00_abstract}

\input{01_intro}
\input{02_related}

\input{03_method}

\input{04_experiment}
\input{10_conclusion}

\noindent\textbf{Acknowledgment.} This research was partially supported by the National Natural Science Foundation of China under Grant 32441114, 41971397, 61731022, and the project Y1H103101A.

{\small
\bibliographystyle{ieeenat_fullname}
\bibliography{11_references}
}

\ifarxiv \clearpage \appendix \input{12_appendix} \fi

\end{document}

%% file: _constants.tex
\def\paperID{4805}
\def\confName{ICCV}
\def\confYear{2025}

\def\paperTitle{Enpowering Your Pansharpening Models with Generalizability: \\ Unified Distribution is All You Need}

\def\authorBlock{
    Yongchuan Cui \textsuperscript{1, 2}\qquad
    Peng Liu \textsuperscript{1, 2,} \footnotemark[1] \qquad 
    Hui Zhang \textsuperscript{3,} \footnotemark[1] \qquad \\
    \textsuperscript{1} Aerospace Information Research Institute, Chinese Academy of Sciences, 100094 Beijing, China \\
    \textsuperscript{2} School of Electronic, Electrical and Communication Engineering, \\ University of Chinese Academy of Sciences, 101408 Beijing, China \\
    \textsuperscript{3} School of Engineering Medicine, Beihang University, 100191 Beijing, China \\
    {\tt\small yongchuancui@gmail.com, liupeng202303@aircas.ac.cn, hui.zhang@buaa.edu.cn}
    \vspace{-1em}
}

\newif\ifreview 
\newif\ifarxiv 
\newif\ifcamera \newcommand{\cameraready}{\cameratrue}
\newif\ifrebuttal 

%% file: cvpr_header.tex
\ifreview \usepackage[review]{cvpr} \fi
\ifarxiv \usepackage[pagenumbers]{cvpr} \fi
\ifrebuttal \usepackage[rebuttal]{cvpr} \fi
\ifcamera \usepackage{cvpr} \fi

\input{_macros}  % Add packages to _macros.tex

\usepackage{xr-hyper}

\makeatletter
\newcommand*{\addFileDependency}[1]{
  \typeout{(#1)}
  \@addtofilelist{#1}
  \IfFileExists{#1}{}{\typeout{No file #1.}}
}

\makeatother
\newcommand*{\myexternaldocument}[1]{
    \externaldocument{#1}
    \addFileDependency{#1.tex}
    \addFileDependency{#1.aux}
}

\definecolor{cvprblue}{rgb}{0.21,0.49,0.74}
\usepackage[pagebackref,breaklinks,colorlinks,allcolors=cvprblue]{hyperref}
\usepackage[capitalize]{cleveref}
\crefname{section}{Sec.}{Secs.}
\crefname{table}{Table}{Tables}
\crefname{figure}{Fig.}{Figs.}

\ifarxiv \crefname{appendix}{App.}{Apps.}
\else \crefname{appendix}{Suppl.}{Suppls.} \fi

%% file: _macros.tex
\usepackage{graphicx}	
\usepackage{amsmath}	
\usepackage{amssymb}	
\usepackage{booktabs}
\usepackage{times}
\usepackage{microtype}
\usepackage{epsfig}
\usepackage{caption}
\usepackage{float}
\usepackage{placeins}
\usepackage{color, colortbl}
\usepackage{stfloats}
\usepackage{enumitem}
\usepackage{tabularx}
\usepackage{xstring}
\usepackage{multirow}
\usepackage{xspace}
\usepackage{url}
\usepackage{subcaption}
\usepackage{xcolor}
\usepackage[hang,flushmargin]{footmisc}

\ifcamera \usepackage[accsupp]{axessibility} \fi

% ...

\newcommand{\nbf}[1]{{\noindent \textbf{#1.}}}

\ifarxiv  \fi

\newcommand{\R}[1]{{%
    \textbf{%
        \ifstrequal{#1}{1}{\textcolor{red}{R#1}}{%
        \ifstrequal{#1}{2}{\textcolor{blue}{R#1}}{%
        \ifstrequal{#1}{3}{\textcolor{magenta}{R#1}}{%
        \ifstrequal{#1}{4}{\textcolor{teal}{R#1}}{%
                           \textcolor{cyan}{R#1}%
        }}}}%
    }%
}}

\usepackage{fontawesome5}
\usepackage{amsthm}
\usepackage{algorithm}      % 算法环境支持
\usepackage{algpseudocode}  % 提供 algorithmicx 的伪代码语法
\usepackage{caption}  % 提供 algorithmicx 的伪代码语法
\usepackage{graphicx}
\usepackage{bigstrut}
\usepackage{subcaption}
\usepackage{float}

\usepackage{autobreak}

\newcommand\Algphase[1]{%
\Statex\hspace*{-6mm}\textbf{#1}%
}

%% file: 00_abstract.tex
\begin{abstract}
Existing deep learning-based models for remote sensing pansharpening exhibit exceptional performance on training datasets. However, due to sensor-specific characteristics and varying imaging conditions, these models suffer from substantial performance degradation when applied to unseen satellite data, lacking generalizability and thus limiting their applicability. We argue that the performance drops stem primarily from distributional discrepancies from different sources and the key to addressing this challenge lies in bridging the gap between training and testing distributions. To validate the idea and further achieve a “train once, deploy forever” capability, this paper introduces a novel and intuitive approach to enpower any pansharpening models with generalizability by employing a unified distribution strategy (UniPAN). Specifically, we construct a distribution transformation function that normalizes the pixels sampled from different sources to conform to an identical distribution. The deep models are trained on the transformed domain, and during testing on new datasets, the new data are also transformed to match the training distribution. UniPAN aims to train and test the model on a unified and consistent distribution, thereby enhancing its generalizability. Extensive experiments validate the efficacy of UniPAN, demonstrating its potential to significantly enhance the performance of deep pansharpening models across diverse satellite sensors. Codes: \url{https://github.com/yc-cui/UniPAN}.
\end{abstract}

%% file: 01_intro.tex
\section{Introduction}

\label{sec:intro}

Due to the inherent limitations of sensor imaging technology, advanced remote sensing satellites such as IKONOS, QuickBird, GaoFen-2, and WorldView-3, necessitate a compromise between the acquisition of extensive spectral information and maintaining fine spatial details in their captured imagery~\cite{surv1,surv2,surv4}. Consequently, these satellites typically capture either high spatial resolution single-band panchromatic (PAN) images or low spatial resolution multispectral (LRMS) images. Pansharpening is a crucial technique in remote sensing image processing, aiming to combine PAN and LRMS images to generate fused images with high spatial resolution and abundant spectra (HRMS). 

\input{figs/tSNE.tex}

Traditional pansharpening methods have been extensively studied over the past few decades. However, these methods may introduce spectral distortions, require significant computational resources, and are sensitive to the quality of input data~\cite{surv1,surv2,surv3,surv4}. In contrast, deep learning (DL)-based methods have shown great potential in pansharpening. Notably, various innovative architectures have emerged, starting from early convolutional models~\cite{PNN,PanNet,MSDCNN,FusionNet}, progressing to Transformer-based approaches~\cite{transformer,ViT,meng2022vision,BDT,PSRT,Hypertransformer,su2022transformer}, then generative methods~\cite{Pan-GAN,PanDiff,PSGAN,rui2024unsupervised,xing2024crossdiff}, and continuing with the emergence of the latest novel designs~\cite{PanMamba,FusionMamba,zhao2024supervised,ZSL,ZS-Pan}. The networks are evolving towards larger parameters capacity, stronger representation ability, and increasingly complex and diverse structures. 

However, as deep pansharpening thrives, the overly powerful fitting capabilities of these networks have concurrently presented several ongoing concerns, \eg generalizability, interpretability, over-reliance on high-quality data, \etc. This paper specifically delves into the widely criticized issue of generalizability. Specifically, although deep models achieve excellent performance on training datasets, they often fail to generalize well to new data due to distribution shift such as different noise levels, varying spectral signatures and sensor conditioning characteristics. While models can achieve satisfactory performances on new datasets through retraining or fine-tuning, the process is time-consuming and fine-tuning on every new dataset is impractical and cumbersome. The ideal scenario would involve developing a pansharpening model that can be trained once and deployed effectively across diverse datasets and sensors, thereby providing a \textit{train once, deploy forever} capability.

To this end, this paper introduces a novel approach that aims to enpower any pansharpening models with generalizability by employing a unified distribution strategy (dubbed as \textbf{UniPAN}). The proposed method aims to bridge the gap between training and testing distributions, thereby improving the model's ability to generalize to unseen data. The core idea is to normalize the distribution of spatial pixels from the training dataset to conform to a predefined distribution (\eg Gaussian distribution) and deep models are then trained on the transformed domain. This process is model-agnostic and requires no additional trainable parameters. When generalizing to new data, testing datasets are also transformed to match the training distribution, ensuring that the models are evaluated under the same distributional assumptions. \cref{fig:intro-cmp} presents the cross-satellite generalization performance of a well-known pansharpening model FusionNet~\cite{FusionNet} on NBUPansharpRSData~\cite{surv2} trained w/ and w/o UniPAN. The model utilizing UniPAN demonstrates a significant improvement on full-resolution metrics, which validates the strong generalization ability for real-world scenes. To further elucidate the efficacy of UniPAN, \cref{fig:tSNE} presents tSNE~\cite{tSNE} dimensionality reduction clustering of multispectral images from diverse satellites using NBUPansharpRSData~\cite{surv2} w/ and w/o UniPAN. The pre-UniPAN visualization reveals distinct clustering patterns, with data points from different satellites forming separate groups, indicative of substantial inter-satellite distributional disparities. In contrast, following the application of UniPAN, a notable transformation is observed: the previously separated clusters become substantially intermingled, demonstrating a significant reduction in distributional discrepancies across different satellite datasets (\eg the \textcolor{Fuchsia}{purple} scatters of QuickBird). This visual evidence underscores UniPAN's capability to harmonize heterogeneous satellite data, effectively creating a more cohesive and unified representation.

The contributions of this paper could be summarized as follows:

\begin{itemize}
    \item We propose UniPAN, a model-agnostic approach that employs a unified distribution strategy to enpower any deep pansharpening models with generalizability, achieving a \textit{train once, deploy forever} capability. This strategy provides two advantages: 1) \textit{Architectural independence}: Compatible with any existing or future network architecture without structural modifications. 2) \textit{Training efficiency}: Eliminates the need for retraining or fine-tuning when adapting to new sensors. 
    \item UniPAN is implemented via inverse transform sampling, and we prove its equivalence to the optimal solution of optimal transport in the one-dimensional case. UniPAN requires no additional trainable parameters and only needs to be fitted once before training begins, introducing almost no additional computational overhead. It serves as a viable alternative to vanilla data normalization techniques such as min-max and $z$-score normalization, offering a more robust and unified approach to data distribution alignment.
    \item We conduct extensive experiments using a variety of deep pansharpening models across numerous satellites. The simple yet intuitive and effective strategy provides a universal solution to the generalization challenge in DL-based pansharpening and experimental results demonstrate its effectiveness, showcasing significant improvements in cross-dataset generalization across a wide range of models and satellite sensors.
\end{itemize}

%% file: figs/tSNE.tex
% Use figure* for multi-column figure
% \begin{figure*}[tp]
%     \centering
%     \includegraphics[width=\linewidth]{figs/tSNE1.pdf}
%     \caption{Caption.}
%     \label{fig:template}
% \end{figure*}

\begin{figure}[t]
    \centering
    \begin{subfigure}[b]{0.325\linewidth}
        \includegraphics[width=\linewidth]{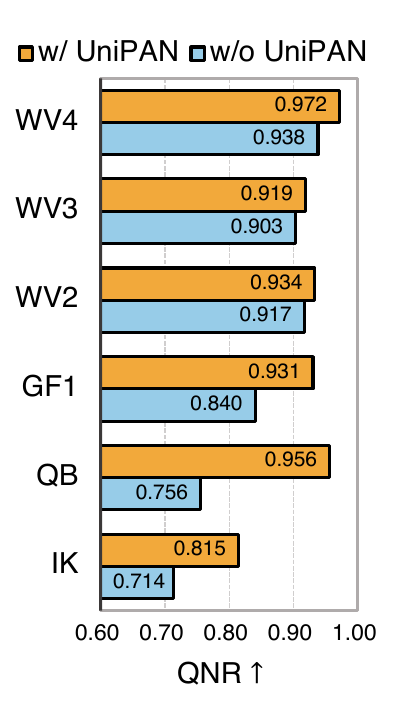}
    \end{subfigure}
    \hfill
    \begin{subfigure}[b]{0.325\linewidth}
        \includegraphics[width=\linewidth]{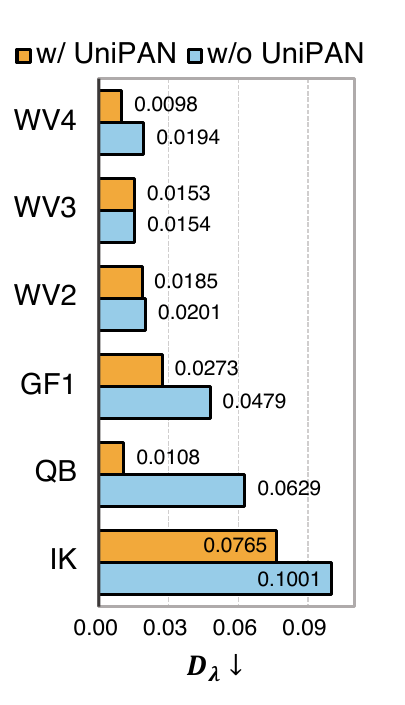}
    \end{subfigure}
    \hfill
    \begin{subfigure}[b]{0.325\linewidth}
        \includegraphics[width=\linewidth]{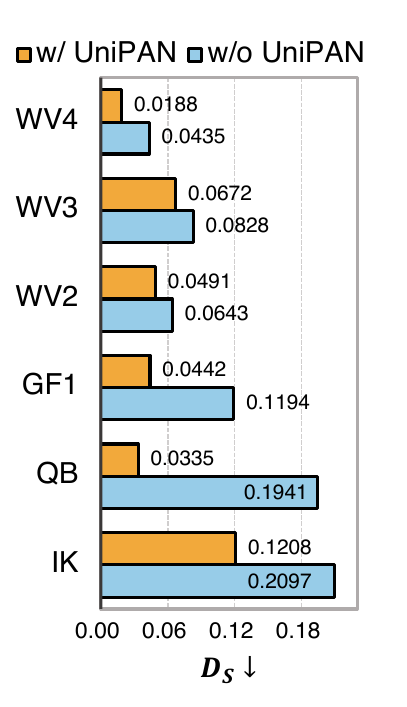}
    \end{subfigure}
    \caption{Comparison of generalization metrics w/ and w/o UniPAN trained on WorldView-3 (WV3) using FusionNet~\cite{FusionNet}.}
    \label{fig:intro-cmp}
\end{figure}

% \caption{Comparison of generalization metrics w/ and w/o UniPAN using FusionNet~\cite{FusionNet}, which was trained on WorldView-3 (WV3) imagery and evaluated on various satellites, including WorldView-4 (WV4), WV3, WorldView-2 (WV2), GaoFen-1 (GF1), QuickBird (QB), and IKONOS (IK).}

% Use figure* for multi-column figure
\begin{figure}[tp]
    \centering
    \vspace{-2mm}
    \includegraphics[width=\linewidth]{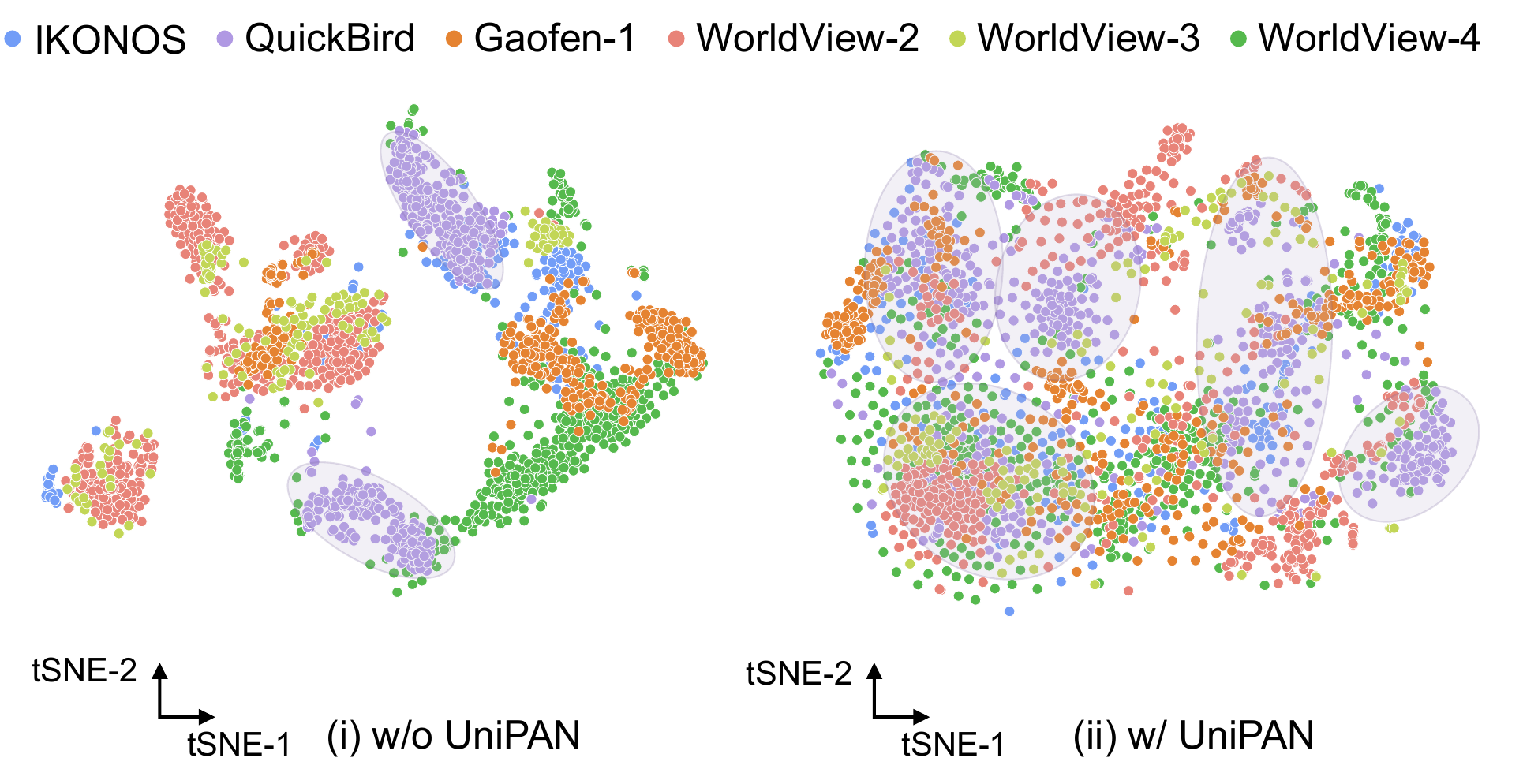}
    \caption{Comparison of tSNE visualization w/ and w/o UniPAN.}
    \vspace{-5mm}
    \label{fig:tSNE}
\end{figure}

% \caption{Comparison of tSNE visualization w/ and w/o UniPAN. After applying UniPAN, all satellites demonstrate more consistent spectral characteristics and clustering patterns are intermingled. Refer to \cref{fig:kde} in \cref{sec:visual} for a more comprehensive comparison of pairwise distribution between different sensors.}

%% file: 02_related.tex
\section{Related Work}
\label{sec:related}

\subsection{Pansharpening Methods}

\nbf{Traditional Methods} Traditional pansharpening methods can be broadly categorized into three paradigms: component substitution (CS), multi-resolution analysis (MRA), and variational optimization (VO)~\cite{surv1,surv2,surv3,surv4}. CS-based methods~\cite{GS,GSA,PCA1,PCA2,IHS1,IHS2} operate by substituting the spatial component of the LRMS image with the PAN image. Representative algorithms include Gram-Schmidt (GS)~\cite{GS,GSA}, principal component analysis (PCA)~\cite{PCA1,PCA2}, and intensity-hue-saturation (IHS)~\cite{IHS1,IHS2}. While computationally efficient, CS methods often introduce spectral distortions due to mismatches between PAN and substituted components. MRA-based methods~\cite{Garzelli2002,Choi2005,MTF-GLP1,MTF-GLP2,MTF-GLP-HPM,SFIM} employ multi-scale decomposition tools (\eg wavelets~\cite{Garzelli2002}, contourlets~\cite{Choi2005}) to inject high-frequency details from PAN into LRMS. Typical implementations include MTF-GLP~\cite{MTF-GLP1,MTF-GLP2,MTF-GLP-HPM} and SFIM~\cite{SFIM}. These methods generally preserve spectral fidelity better than CS approaches but may exhibit limited spatial enhancement. VO-based approaches~\cite{FE-HPM,10330008,Ballester2006,Palsson2014,10218368,VO+Net} formulate pansharpening as an optimization problem with handcrafted priors, such as the variational wavelet-based methods~\cite{Ballester2006} and Bayesian approaches~\cite{Palsson2014}. Despite their mathematical elegance, traditional methods often require manual parameter tuning and struggle with complex spatial-spectral relationships~\cite{surv4,surv1}.

\nbf{Deep Learning-Based Methods} The advent of deep learning (DL) have revolutionized pansharpening by enabling end-to-end learning of spatial-spectral relationships. Early pioneering convolutional neural networks (CNNs) such as PNN~\cite{PNN}, PanNet~\cite{PanNet}, and MSDCNN~\cite{MSDCNN} established the feasibility of learning fusion rules from a mount of simulated data. Subsequent innovations introduced multi-scale architectures~\cite{DMR-Pan,CMLNet,LPPN}, attention mechanisms~\cite{CADUI,CMINet}, and kernel custimizing~\cite{CANConv,LGPConv,SpanConv,LAGConv,PreMix}, \etc. The CNN-based methods demonstrated superior performance over traditional approaches but faced limitations in modeling long-range dependencies. In contrast, Transformer~\cite{transformer,ViT,swin} addressed this limitation with self-attention mechanisms. Although various Transformer architectures specifically designed for pansharpening~\cite{meng2022vision,BDT,PSRT,Hypertransformer,su2022transformer} have achieved impressive performance, they face criticism for quadratic computational complexity \wrt sequence length~\cite{PEMAE}. To address this limitation, recent studies have explored linear-complexity models~\cite{PEMAE,PanMamba,zhao2024supervised,FusionMamba} that maintain performance while improving efficiency. To further enhance fidelity preservation, generative models like generative adversarial networks (GANs)~\cite{cui2024reconstruction,Pan-GAN,PSGAN,Mun-GAN,UPanGAN} and diffusion models~\cite{SSDiff,PanDiff,rui2024unsupervised,xing2024empower,xing2024crossdiff} have emerged as promising solutions. Beyond conventional configurations, contemporary research has expanded into more challenging frontiers to address practical constraints, including zero-shot learning paradigms~\cite{ZS-Pan,PSDip,ZeroSharpen,ZSL}, frequency domain integration~\cite{FeINFN,SFIIN,SFIIN++,WINet,FAME-Net}, and interpretable fusion mechanisms~\cite{GPPNN,DISPNet,MDCUN,BUGPan}.

\subsection{Generalization in DL-Based Pansharpening}
\label{subsec:generalization }

Despite remarkable progress in DL-based pansharpening, generalization capability remains a persistent challenge. The issue manifests conspicuously in two critical scenarios: 1) Models trained under the Wald protocol~\cite{wald} using spatially degraded data often fail to maintain performance when applied to full-resolution inference, frequently exhibiting severe spectral distortions. 2) Networks optimized for specific satellite sensors demonstrate substantial performance degradation when deployed on unseen satellite platforms due to distribution shifts in spectral signatures and sensor characteristics. Existing approaches to address these challenges primarily focus on unsupervised learning~\cite{Pan-GAN,Z-PNN,Lambda-PNN}, generalized module design~\cite{xing2024empower}, or advanced data augmentation~\cite{10082993}. However, such methods typically require either meticulous architectural redesign or computationally intensive retraining/fine-tuning procedures. These requirements pose significant practical limitations for real-world deployment.

In contrast to model-centric solutions, our approach adopts a fundamentally different perspective by addressing the generalization challenge through data-space intervention. The proposed UniPAN framework introduces a model-agnostic preprocessing technique that operates on raw sensor data. Compared to traditional linear data normalization such as min-max scaling or $z$-score standardization, our method employs a non-linear mapping to project the input data onto a unified distribution, effectively harmonizing the distribution mismatches from all sensors.

%% file: 03_method.tex
\section{Method}
\label{sec:method}

\subsection{Distribution Transformation}

The core challenge in cross-domain pansharpening stems from the distribution discrepancy between source (training) and target (testing) domains. Let $\mathcal{X}_S$ denote the source domain (training data) and $\mathcal{X}_T$ the target domain (testing data). Our objective is to learn a transformation $\mathcal{T}: \mathcal{X} \rightarrow \mathcal{Z}$ that maps both domains to a shared space $\mathcal{Z}$ with unified distribution $\rho_{t}$, \eg standard normal distribution.

\nbf{Inverse Transform Sampling} Given a univariate r.v. $X \sim \rho_{s}$, we seek a transformation $\mathcal{T}: \mathbb{R} \rightarrow \mathbb{R}$ such that $Z = \mathcal{T}(X) \sim \rho_{t}$. The inverse transform sampling theorem provides a theoretically grounded solution:
\begin{equation}
    \label{eq:ITS}
    \mathcal{T}(x) = F_T^{-1}(F_S(x)),
    \end{equation}
where $F_S(x) =\rho_s(-\infty,x]$ is the cumulative distribution function (CDF) of $X$, and $F^{-1}_T(u) = \inf\{z \in \mathbb{R} : F_T(z) \geq u\}$ is the inverse CDF of the target distribution. By the universality of the uniform, $U = F_S(X)$ follows a uniform distribution, and $Z = F^{-1}_T(U)$ inherits $\rho_{t}$ exactly when $F_S$ is continuous. To provide more intuitively analyze of the algorithm's operational process and efficiency, we establish a fundamental connection between inverse transform sampling and the well-studied optimal transport theory~\cite{monge1781memoire,COTFNT}.

\nbf{Optimal Transport Perspective} Given two probability distributions $\rho_s$ (source domain) and $\rho_t$ (target domain) defined on $\mathbb{R}^n$ and $\mathbb{R}^m$ respectively, we consider the Monge problem~\cite{monge1781memoire,COTFNT}: 
\begin{equation}
    \label{eq:Monge}
    \mathcal{C}(\rho_s,\rho_t) \triangleq \min_{\scriptsize{\begin{array}{c}T:\mathbb{R}^n\to\mathbb{R}^m\\T_\sharp\rho_s=\rho_t\end{array}}}\int_{\mathbb{R}^n}c(x,T(x))\rho_s(x)\mathrm{~}dx,
\end{equation}
where $c:\Omega_s\times\Omega_t\to\mathbb{R}^+$ is a cost function and for any measurable Borel set $E \in \Omega_s$, $T:\Omega_s\to\Omega_t$ transfers measures from one space $\Omega_s$ to another space $\Omega_t$,
\begin{equation}
    T_\sharp\rho_s(E)=\rho_s(T^{-1}(E)),
\end{equation}
here $T_\sharp$ is the pushforward operator. The Monge problem seeks a transport plan $T$ from $\rho_s$ to $\rho_t$ that minimizes the transportation cost. The optimal $T^{*}$ is also known as the Monge map of \cref{eq:Monge}.

Let $F_S(x)=\rho_s(-\infty,x]$ and $F_T(z)=\rho_t(-\infty,z]$ denote the CDFs of the source and target distributions respectively. In the univariate setting ($n=m=1, \Omega_s = \Omega_t = \mathbb{R}$), when the source measure $\rho_s$ is continuous and the cost function takes the form $c(x,z)=h(|x-z|)$ with $h$ being strictly convex, by cyclical monotonicity~\cite{villani2008optimal}, the Monge map $T^*$ satisfies:
\begin{equation}
    F_S(x) = F_T(T^*(x)),
\end{equation}
yielding the explicit expression:
\begin{equation}
    T^*(x) = F_T^{-1}(F_S(x)).
\end{equation}

This remarkable result coincides exactly with the inverse transform sampling formula \cref{eq:ITS}, \ie $T^* = \mathcal{T}$. The transport cost in this setting~\cite{kolouri2019generalizedslicedwassersteindistances,ramdas2015wassersteinsampletestingrelated,Santambrogio2015} can be computed as
\begin{equation}
    \mathcal{C}(\rho_s,\rho_t) = \int_0^1 c(F_S^{-1}(u),F_T^{-1}(u))du.
\end{equation}

For the $L_1$ cost function $c(x,z)=|x-z|$ , this corresponds to the $W_1$ Wasserstein distance~\cite{Santambrogio2015,7974883} between the distributions. In the one-dimensional (1D) scenario, UniPAN operates independently on spatial pixels within each spectral band. This implementation can be achieved through interpolation-based inverse transform sampling without framing it as an optimization problem. However, given the inherent correlations between spectral bands and interdependencies among spatial patches, multidimensional extension of this framework may be promising, albeit with increased implementation complexity. The theoretical derivation establishing connections with optimal transport theory serves to facilitate future generalization of UniPAN to multidimensional configurations. Building upon the theoretical foundation of distribution transformation through optimal transport, in the following, we present a practical implementation of 1D UniPAN.

\subsection{UniPAN Implementation}

We present a practical implementation for 1D finite discrete distributions of UniPAN through a quantile-based distribution transformation algorithm. 

\nbf{Finite Discrete Distribution Transformation} Given pixel values $X = \{x_1, x_2, ..., x_n\}$ sampled from a spectral band, we first compute $m$ empirical quantiles $Q = \{q_0, ..., q_{m-1}\}$ where:
\begin{equation}
    \label{eq:quantiles}
    q_i = \inf\left\{x \in X \big| P(X \leq x) \geq \frac{i}{m-1}\right\}, \small{i=\{0, 1, ..., m-1\}}.
\end{equation}

These quantiles discretize the empirical cumulative distribution function (eCDF) of the source data. The number of quantiles $m$ controls the approximation fidelity.

\input{tables/alg1.tex}

Given any data point $x$ to be transformed, the next step involves determining its value in the eCDF through interpolation. However, when the source distribution contains repeated values or sparse regions, quantile estimation may yield non-unique or overlapping quantile boundaries. This can result in abrupt jumps in the eCDF, thereby violating the properties of monotonicity and smoothness. Therefore, we use bidirectional interpolation to estimate a robust eCDF value of $x$:
\begin{equation}
    \label{eq:ux}
    \begin{aligned}
u(x) &= u_{\uparrow}(x) / 2 + u_{\downarrow}(x) / 2 \\
&= \texttt{interp}(x, Q, U) \, / \, 2 \, + \\  & \quad  \left(1  - \texttt{interp}\left(-x, -Q^{\text{rev}}, -U^{\text{rev}}\right)\right) \,/ \,2,
\end{aligned}
\end{equation}
where $U = \{0, \frac{1}{m-1}, ..., 1\}$, $Q^{\text{rev}}$ is the reversed quantiles. The function $\texttt{interp}$ represents a linear interpolation between the quantile point $Q$ and the reference point $U$, with the interpolation formula given by:
\begin{equation}
    \label{eq:interp}
  \texttt{interp}(x, Q, U) = 
  \begin{cases}
    U[0], & x \leq Q[0] \\
    f(x, i, Q, U),& Q[i] < x \leq Q[i+1]\\
    U[m-1], & x > Q[m-1]
  \end{cases}
\end{equation}
\noindent where $f$ computes the $Q$ to $U$ mapping,
\begin{equation}
f(x, i, Q, U) = U[i] + \frac{x - Q[i]}{Q[i+1] - Q[i]} (U[i+1] - U[i])
\end{equation}

The last step is to apply inverse CDF of target distribution,
\begin{equation}
    \label{eq:final_trans}
    \mathcal{T}(x) = F_T^{-1}(u(x))
\end{equation}

\input{tables/cmp-all-avg.tex}

\nbf{Training and Testing Procedure} The training and testing procedures of UniPAN are described in~\cref{alg:unipan}. Here, $L\!\uparrow$ denotes the upsampled LRMS image using bicubic interpolation to match the size of the PAN image. The model $\mathcal{M}$ is trained to predict the residual of LRMS with an $l_1$ loss function that minimizes the discrepancy between the predicted HRMS image and the ground truth.

As for the target distribution $\rho_t$, we test two widely used distributions, \ie standard normal distribution $\mathcal{N}(0,1)$ and standard uniform distribution $\mathcal{U}(0,1)$, the transformation $\mathcal{T}(x)$ becomes:
\begin{equation}
    \mathcal{T}_{\text{n}}(x) = \Phi^{-1}(F_S(x)), \quad \mathcal{T}_{\text{u}}(x) = F_S(x),
\end{equation}
where $\Phi^{-1}$ denotes the inverse CDF of standard normal distribution. Since the inverse CDF of standard uniform distribution is an identical mapping, $\mathcal{T}_{\text{u}}$ becomes the estimated empirical distribution itself. The standard normal distribution transformation non-linearly warps the input space to match Gaussian statistics, while the uniform transformation directly maps pixel intensities to their empirical quantiles.

% Figure \cref{fig:dist_trans} visualizes the transformation effects. The original PAN histogram exhibits a bimodal distribution (Fig. \cref{fig:dist_trans}a). After $\mathcal{T}_{\text{unif}}$ (Fig. \cref{fig:dist_trans}b), the histogram becomes approximately uniform while preserving the original value ordering. The $\mathcal{T}_{\text{norm}}$ transformation (Fig. \cref{fig:dist_trans}c) produces a Gaussian-like distribution, with the eCDF closely matching the standard normal CDF. Both transformations maintain the spatial structure of input images while achieving distributional alignment.

%% file: tables/alg1.tex
\begin{algorithm}[t]
    \caption{UniPAN Training and Testing Procedure}
    \label{alg:unipan}
    \begin{algorithmic}[1]
    \small
    \Algphase{Training Phase}
    \Require Training dataset $\mathcal{D}_{\text{train}} = \{L_i, P_i, H_i\}_{i=1}^{N}$, inverse CDF $F_T^{-1}$ of target distribution $\rho_t$, model $\mathcal{M}$
    \Ensure Trained model $M$
    
    \State \textbf{Initialize} transformed LRMS $\tilde{L}$ and PAN $\tilde{P}$
    \For{each band $b$ of LRMS $\{L_i\}_{i=1}^{N}$  \& PAN $\{P_i\}_{i=1}^{N}$}
        \State Sample $n$ pixels $X_b \gets \{x_1,\dots,x_n\}$ from band $b$
        \State Compute $m$ quantiles $Q = \{q_0,\dots,q_{m-1}\}$  \hfill $\triangleright$ \cref{eq:quantiles}
        \State Linspace uniform CDF $U = \{0, \frac{1}{m-1}, ..., 1\}$
        \For{each pixel $x$ in band $b$}
            \State Forward and backward interp by $u_{\uparrow}$ and $u_{\downarrow}$ \hfill $\triangleright$ \cref{eq:interp}
            \State Compute eCDF value $u(x) \gets (u_{\uparrow} + u_{\downarrow}) \,/\,2$ \hfill $\triangleright$ \cref{eq:ux}
            \State Transform $x$ to $z \gets F_T^{-1}(u(x))$ and assign $z$ to corresponding band of $\tilde{L}$ and $\tilde{P}$ \hspace{27mm} $\triangleright$ \cref{eq:final_trans} 
        \EndFor
    \EndFor
    \While{not converged}
        \State Train $\mathcal{M}$ with $\mathcal{L} = \|H - \mathcal{M}(\tilde{L}, \tilde{P}) + L\!\uparrow\|_1$
    \EndWhile
    \State \Return Trained model $\mathcal{M}$
    \vspace{-.3\baselineskip}
    \Statex\hspace*{\dimexpr-\algorithmicindent-2pt\relax}\rule{0.47\textwidth}{0.4pt}%
    \Algphase{Testing Phase}
    \Require Testing dataset $\mathcal{D}_{\text{test}} = \{L_i, P_i\}_{i=1}^{N}$, inverse CDF $F_T^{-1}$ of target distribution $\rho_t$, trained model $\mathcal{M}$
    \Ensure Pansharpening result $\hat{H}^{\text{test}}$
    
    \For{each band $b$ of LRMS \& PAN}
        \State Transform $L, P$ to $\tilde{L}, \tilde{P}$ using the training transformation
    \EndFor
    
    \State Generate $\hat{H}^{\text{test}} \gets \mathcal{M}(\tilde{L}, \tilde{P})+ L\!\uparrow $
    \State \Return Fused result $\hat{H}^{\text{test}}$
    
    \end{algorithmic}
    \end{algorithm}

%% file: tables/cmp-all-avg.tex
\begin{table*}[htbp]
  \centering  
  \vspace{-3mm}
  \resizebox{\linewidth}{!}{
  \begin{tabular}{lrrrrrrrrrrrr}
  \toprule
  \multicolumn{1}{c}{\multirow{2}[4]{*}{QNR$\uparrow$}} & \multicolumn{6}{c}{$\rho_t=\mathcal{N}(0, 1)$} & \multicolumn{6}{c}{$\rho_t=\mathcal{U}(0, 1)$} \bigstrut[t]\\
  \cmidrule(r){2-7}  \cmidrule(r){8-13}      & \multicolumn{1}{c}{GF1} & \multicolumn{1}{c}{IK} & \multicolumn{1}{c}{QB} & \multicolumn{1}{c}{WV2} & \multicolumn{1}{c}{WV3} & \multicolumn{1}{c}{WV4} & \multicolumn{1}{c}{GF1} & \multicolumn{1}{c}{IK} & \multicolumn{1}{c}{QB} & \multicolumn{1}{c}{WV2} & \multicolumn{1}{c}{WV3} & \multicolumn{1}{c}{WV4} \bigstrut\\
  \hline
  GF1  &\cellcolor{green!4}$0.980(\textcolor{PineGreen}{+0.005})$ & \cellcolor{red!4}$0.946(\textcolor{BrickRed}{-0.010})$ & \cellcolor{green!4}$0.971(\textcolor{PineGreen}{+0.000})$ & \cellcolor{red!4}$0.963(\textcolor{BrickRed}{-0.003})$ & \cellcolor{red!4}$0.964(\textcolor{BrickRed}{-0.004})$ & \cellcolor{red!4}$0.974(\textcolor{BrickRed}{-0.000})$ & \cellcolor{green!4}$0.985(\textcolor{PineGreen}{+0.010})$ & \cellcolor{green!4}$0.961(\textcolor{PineGreen}{+0.004})$ & \cellcolor{green!4}$0.979(\textcolor{PineGreen}{+0.007})$ & \cellcolor{green!4}$0.984(\textcolor{PineGreen}{+0.017})$ & \cellcolor{green!4}$0.982(\textcolor{PineGreen}{+0.013})$ & \cellcolor{green!4}$0.987(\textcolor{PineGreen}{+0.013})$ \bigstrut[t]\\
  IK   &\cellcolor{green!4}$0.950(\textcolor{PineGreen}{+0.005})$ & \cellcolor{green!4}$0.912(\textcolor{PineGreen}{+0.039})$ & \cellcolor{green!4}$0.915(\textcolor{PineGreen}{+0.045})$ & \cellcolor{green!4}$0.924(\textcolor{PineGreen}{+0.011})$ & \cellcolor{green!4}$0.916(\textcolor{PineGreen}{+0.022})$ & \cellcolor{green!4}$0.958(\textcolor{PineGreen}{+0.011})$ & \cellcolor{green!4}$0.956(\textcolor{PineGreen}{+0.011})$ & \cellcolor{green!4}$0.919(\textcolor{PineGreen}{+0.045})$ & \cellcolor{green!4}$0.940(\textcolor{PineGreen}{+0.070})$ & \cellcolor{green!4}$0.950(\textcolor{PineGreen}{+0.037})$ & \cellcolor{green!4}$0.945(\textcolor{PineGreen}{+0.051})$ & \cellcolor{green!4}$0.968(\textcolor{PineGreen}{+0.021})$ \bigstrut[t]\\
  QB   &\cellcolor{green!4}$0.957(\textcolor{PineGreen}{+0.006})$ & \cellcolor{green!4}$0.936(\textcolor{PineGreen}{+0.016})$ & \cellcolor{green!4}$0.931(\textcolor{PineGreen}{+0.047})$ & \cellcolor{green!4}$0.953(\textcolor{PineGreen}{+0.017})$ & \cellcolor{green!4}$0.944(\textcolor{PineGreen}{+0.014})$ & \cellcolor{green!4}$0.970(\textcolor{PineGreen}{+0.016})$ & \cellcolor{green!4}$0.965(\textcolor{PineGreen}{+0.014})$ & \cellcolor{green!4}$0.944(\textcolor{PineGreen}{+0.024})$ & \cellcolor{green!4}$0.966(\textcolor{PineGreen}{+0.082})$ & \cellcolor{green!4}$0.969(\textcolor{PineGreen}{+0.032})$ & \cellcolor{green!4}$0.969(\textcolor{PineGreen}{+0.038})$ & \cellcolor{green!4}$0.970(\textcolor{PineGreen}{+0.017})$ \bigstrut[t]\\
  WV2  &\cellcolor{green!4}$0.880(\textcolor{PineGreen}{+0.026})$ & \cellcolor{green!4}$0.811(\textcolor{PineGreen}{+0.029})$ & \cellcolor{green!4}$0.832(\textcolor{PineGreen}{+0.062})$ & \cellcolor{green!4}$0.946(\textcolor{PineGreen}{+0.008})$ & \cellcolor{green!4}$0.916(\textcolor{PineGreen}{+0.012})$ & \cellcolor{green!4}$0.951(\textcolor{PineGreen}{+0.012})$ & \cellcolor{green!4}$0.872(\textcolor{PineGreen}{+0.018})$ & \cellcolor{red!4}$0.775(\textcolor{BrickRed}{-0.006})$ & \cellcolor{green!4}$0.825(\textcolor{PineGreen}{+0.055})$ & \cellcolor{green!4}$0.944(\textcolor{PineGreen}{+0.006})$ & \cellcolor{green!4}$0.914(\textcolor{PineGreen}{+0.011})$ & \cellcolor{green!4}$0.940(\textcolor{PineGreen}{+0.001})$ \bigstrut[t]\\
  WV3  &\cellcolor{red!4}$0.862(\textcolor{BrickRed}{-0.006})$ & \cellcolor{green!4}$0.804(\textcolor{PineGreen}{+0.035})$ & \cellcolor{green!4}$0.795(\textcolor{PineGreen}{+0.067})$ & \cellcolor{red!4}$0.921(\textcolor{BrickRed}{-0.011})$ & \cellcolor{green!4}$0.919(\textcolor{PineGreen}{+0.011})$ & \cellcolor{green!4}$0.950(\textcolor{PineGreen}{+0.001})$ & \cellcolor{green!4}$0.880(\textcolor{PineGreen}{+0.011})$ & \cellcolor{green!4}$0.805(\textcolor{PineGreen}{+0.036})$ & \cellcolor{green!4}$0.813(\textcolor{PineGreen}{+0.084})$ & \cellcolor{green!4}$0.943(\textcolor{PineGreen}{+0.010})$ & \cellcolor{green!4}$0.924(\textcolor{PineGreen}{+0.016})$ & \cellcolor{green!4}$0.952(\textcolor{PineGreen}{+0.003})$ \bigstrut[t]\\
  WV4  &\cellcolor{green!4}$0.924(\textcolor{PineGreen}{+0.009})$ & \cellcolor{green!4}$0.898(\textcolor{PineGreen}{+0.045})$ & \cellcolor{green!4}$0.875(\textcolor{PineGreen}{+0.036})$ & \cellcolor{red!4}$0.948(\textcolor{BrickRed}{-0.000})$ & \cellcolor{green!4}$0.929(\textcolor{PineGreen}{+0.003})$ & \cellcolor{green!4}$0.972(\textcolor{PineGreen}{+0.006})$ & \cellcolor{green!4}$0.939(\textcolor{PineGreen}{+0.024})$ & \cellcolor{green!4}$0.889(\textcolor{PineGreen}{+0.035})$ & \cellcolor{green!4}$0.903(\textcolor{PineGreen}{+0.064})$ & \cellcolor{green!4}$0.952(\textcolor{PineGreen}{+0.002})$ & \cellcolor{green!4}$0.935(\textcolor{PineGreen}{+0.009})$ & \cellcolor{green!4}$0.973(\textcolor{PineGreen}{+0.007})$ \bigstrut[t]\\
  \hline
  \multicolumn{1}{c}{\multirow{2}[4]{*}{$D_{\lambda}\downarrow$}} & \multicolumn{6}{c}{$\rho_t=\mathcal{N}(0, 1)$} & \multicolumn{6}{c}{$\rho_t=\mathcal{U}(0, 1)$} \bigstrut[t]\\
  \cmidrule(r){2-7}  \cmidrule(r){8-13}      & \multicolumn{1}{c}{GF1} & \multicolumn{1}{c}{IK} & \multicolumn{1}{c}{QB} & \multicolumn{1}{c}{WV2} & \multicolumn{1}{c}{WV3} & \multicolumn{1}{c}{WV4} & \multicolumn{1}{c}{GF1} & \multicolumn{1}{c}{IK} & \multicolumn{1}{c}{QB} & \multicolumn{1}{c}{WV2} & \multicolumn{1}{c}{WV3} & \multicolumn{1}{c}{WV4} \bigstrut\\
  \hline
  GF1  &\cellcolor{green!4}$0.007(\textcolor{PineGreen}{-0.003})$ & \cellcolor{red!4}$0.022(\textcolor{BrickRed}{+0.008})$ & \cellcolor{green!4}$0.005(\textcolor{PineGreen}{-0.000})$ & \cellcolor{red!4}$0.017(\textcolor{BrickRed}{+0.004})$ & \cellcolor{red!4}$0.015(\textcolor{BrickRed}{+0.005})$ & \cellcolor{red!4}$0.011(\textcolor{BrickRed}{+0.000})$ & \cellcolor{green!4}$0.004(\textcolor{PineGreen}{-0.005})$ & \cellcolor{green!4}$0.012(\textcolor{PineGreen}{-0.001})$ & \cellcolor{green!4}$0.004(\textcolor{PineGreen}{-0.001})$ & \cellcolor{green!4}$0.004(\textcolor{PineGreen}{-0.008})$ & \cellcolor{green!4}$0.004(\textcolor{PineGreen}{-0.005})$ & \cellcolor{green!4}$0.004(\textcolor{PineGreen}{-0.005})$ \bigstrut[t]\\
  IK   &\cellcolor{green!4}$0.011(\textcolor{PineGreen}{-0.000})$ & \cellcolor{green!4}$0.020(\textcolor{PineGreen}{-0.015})$ & \cellcolor{green!4}$0.018(\textcolor{PineGreen}{-0.011})$ & \cellcolor{red!4}$0.031(\textcolor{BrickRed}{+0.004})$ & \cellcolor{green!4}$0.022(\textcolor{PineGreen}{-0.003})$ & \cellcolor{green!4}$0.014(\textcolor{PineGreen}{-0.003})$ & \cellcolor{green!4}$0.008(\textcolor{PineGreen}{-0.003})$ & \cellcolor{green!4}$0.020(\textcolor{PineGreen}{-0.015})$ & \cellcolor{green!4}$0.013(\textcolor{PineGreen}{-0.016})$ & \cellcolor{green!4}$0.013(\textcolor{PineGreen}{-0.014})$ & \cellcolor{green!4}$0.009(\textcolor{PineGreen}{-0.016})$ & \cellcolor{green!4}$0.010(\textcolor{PineGreen}{-0.008})$ \bigstrut[t]\\
  QB   &\cellcolor{green!4}$0.013(\textcolor{PineGreen}{-0.000})$ & \cellcolor{green!4}$0.011(\textcolor{PineGreen}{-0.004})$ & \cellcolor{green!4}$0.015(\textcolor{PineGreen}{-0.010})$ & \cellcolor{green!4}$0.016(\textcolor{PineGreen}{-0.004})$ & \cellcolor{red!4}$0.014(\textcolor{BrickRed}{+0.000})$ & \cellcolor{green!4}$0.010(\textcolor{PineGreen}{-0.005})$ & \cellcolor{green!4}$0.009(\textcolor{PineGreen}{-0.003})$ & \cellcolor{green!4}$0.009(\textcolor{PineGreen}{-0.006})$ & \cellcolor{green!4}$0.007(\textcolor{PineGreen}{-0.017})$ & \cellcolor{green!4}$0.008(\textcolor{PineGreen}{-0.013})$ & \cellcolor{green!4}$0.006(\textcolor{PineGreen}{-0.007})$ & \cellcolor{green!4}$0.012(\textcolor{PineGreen}{-0.003})$ \bigstrut[t]\\
  WV2  &\cellcolor{green!4}$0.042(\textcolor{PineGreen}{-0.006})$ & \cellcolor{green!4}$0.066(\textcolor{PineGreen}{-0.007})$ & \cellcolor{green!4}$0.037(\textcolor{PineGreen}{-0.019})$ & \cellcolor{green!4}$0.008(\textcolor{PineGreen}{-0.000})$ & \cellcolor{green!4}$0.014(\textcolor{PineGreen}{-0.000})$ & \cellcolor{green!4}$0.016(\textcolor{PineGreen}{-0.002})$ & \cellcolor{green!4}$0.046(\textcolor{PineGreen}{-0.002})$ & \cellcolor{red!4}$0.086(\textcolor{BrickRed}{+0.012})$ & \cellcolor{green!4}$0.048(\textcolor{PineGreen}{-0.008})$ & \cellcolor{red!4}$0.008(\textcolor{BrickRed}{+0.000})$ & \cellcolor{red!4}$0.015(\textcolor{BrickRed}{+0.001})$ & \cellcolor{red!4}$0.022(\textcolor{BrickRed}{+0.002})$ \bigstrut[t]\\
  WV3  &\cellcolor{red!4}$0.050(\textcolor{BrickRed}{+0.010})$ & \cellcolor{green!4}$0.067(\textcolor{PineGreen}{-0.011})$ & \cellcolor{green!4}$0.049(\textcolor{PineGreen}{-0.024})$ & \cellcolor{red!4}$0.027(\textcolor{BrickRed}{+0.011})$ & \cellcolor{red!4}$0.014(\textcolor{BrickRed}{+0.000})$ & \cellcolor{red!4}$0.016(\textcolor{BrickRed}{+0.002})$ & \cellcolor{red!4}$0.042(\textcolor{BrickRed}{+0.002})$ & \cellcolor{green!4}$0.072(\textcolor{PineGreen}{-0.006})$ & \cellcolor{green!4}$0.051(\textcolor{PineGreen}{-0.022})$ & \cellcolor{green!4}$0.013(\textcolor{PineGreen}{-0.002})$ & \cellcolor{green!4}$0.011(\textcolor{PineGreen}{-0.002})$ & \cellcolor{red!4}$0.014(\textcolor{BrickRed}{+0.000})$ \bigstrut[t]\\
  WV4  &\cellcolor{green!4}$0.024(\textcolor{PineGreen}{-0.000})$ & \cellcolor{green!4}$0.031(\textcolor{PineGreen}{-0.012})$ & \cellcolor{green!4}$0.029(\textcolor{PineGreen}{-0.009})$ & \cellcolor{red!4}$0.017(\textcolor{BrickRed}{+0.008})$ & \cellcolor{red!4}$0.018(\textcolor{BrickRed}{+0.008})$ & \cellcolor{green!4}$0.005(\textcolor{PineGreen}{-0.001})$ & \cellcolor{green!4}$0.016(\textcolor{PineGreen}{-0.008})$ & \cellcolor{green!4}$0.032(\textcolor{PineGreen}{-0.011})$ & \cellcolor{green!4}$0.022(\textcolor{PineGreen}{-0.016})$ & \cellcolor{red!4}$0.014(\textcolor{BrickRed}{+0.004})$ & \cellcolor{red!4}$0.015(\textcolor{BrickRed}{+0.005})$ & \cellcolor{green!4}$0.004(\textcolor{PineGreen}{-0.002})$ \bigstrut[t]\\
  \hline
  \multicolumn{1}{c}{\multirow{2}[4]{*}{$D_{S}\downarrow$}} & \multicolumn{6}{c}{$\rho_t=\mathcal{N}(0, 1)$} & \multicolumn{6}{c}{$\rho_t=\mathcal{U}(0, 1)$} \bigstrut[t]\\
  \cmidrule(r){2-7}  \cmidrule(r){8-13}      & \multicolumn{1}{c}{GF1} & \multicolumn{1}{c}{IK} & \multicolumn{1}{c}{QB} & \multicolumn{1}{c}{WV2} & \multicolumn{1}{c}{WV3} & \multicolumn{1}{c}{WV4} & \multicolumn{1}{c}{GF1} & \multicolumn{1}{c}{IK} & \multicolumn{1}{c}{QB} & \multicolumn{1}{c}{WV2} & \multicolumn{1}{c}{WV3} & \multicolumn{1}{c}{WV4} \bigstrut\\
  \hline
  GF1  &\cellcolor{green!4}$0.012(\textcolor{PineGreen}{-0.003})$ & \cellcolor{red!4}$0.032(\textcolor{BrickRed}{+0.003})$ & \cellcolor{red!4}$0.023(\textcolor{BrickRed}{+0.000})$ & \cellcolor{red!4}$0.020(\textcolor{BrickRed}{+0.000})$ & \cellcolor{red!4}$0.021(\textcolor{BrickRed}{+0.000})$ & \cellcolor{green!4}$0.015(\textcolor{PineGreen}{-0.000})$ & \cellcolor{green!4}$0.010(\textcolor{PineGreen}{-0.005})$ & \cellcolor{green!4}$0.026(\textcolor{PineGreen}{-0.002})$ & \cellcolor{green!4}$0.016(\textcolor{PineGreen}{-0.006})$ & \cellcolor{green!4}$0.010(\textcolor{PineGreen}{-0.009})$ & \cellcolor{green!4}$0.013(\textcolor{PineGreen}{-0.007})$ & \cellcolor{green!4}$0.007(\textcolor{PineGreen}{-0.008})$ \bigstrut[t]\\
  IK   &\cellcolor{green!4}$0.039(\textcolor{PineGreen}{-0.005})$ & \cellcolor{green!4}$0.068(\textcolor{PineGreen}{-0.026})$ & \cellcolor{green!4}$0.068(\textcolor{PineGreen}{-0.036})$ & \cellcolor{green!4}$0.046(\textcolor{PineGreen}{-0.014})$ & \cellcolor{green!4}$0.063(\textcolor{PineGreen}{-0.019})$ & \cellcolor{green!4}$0.027(\textcolor{PineGreen}{-0.007})$ & \cellcolor{green!4}$0.035(\textcolor{PineGreen}{-0.008})$ & \cellcolor{green!4}$0.062(\textcolor{PineGreen}{-0.033})$ & \cellcolor{green!4}$0.046(\textcolor{PineGreen}{-0.058})$ & \cellcolor{green!4}$0.037(\textcolor{PineGreen}{-0.024})$ & \cellcolor{green!4}$0.045(\textcolor{PineGreen}{-0.036})$ & \cellcolor{green!4}$0.022(\textcolor{PineGreen}{-0.013})$ \bigstrut[t]\\
  QB   &\cellcolor{green!4}$0.030(\textcolor{PineGreen}{-0.005})$ & \cellcolor{green!4}$0.053(\textcolor{PineGreen}{-0.012})$ & \cellcolor{green!4}$0.054(\textcolor{PineGreen}{-0.039})$ & \cellcolor{green!4}$0.031(\textcolor{PineGreen}{-0.012})$ & \cellcolor{green!4}$0.042(\textcolor{PineGreen}{-0.014})$ & \cellcolor{green!4}$0.020(\textcolor{PineGreen}{-0.012})$ & \cellcolor{green!4}$0.025(\textcolor{PineGreen}{-0.011})$ & \cellcolor{green!4}$0.046(\textcolor{PineGreen}{-0.018})$ & \cellcolor{green!4}$0.026(\textcolor{PineGreen}{-0.067})$ & \cellcolor{green!4}$0.023(\textcolor{PineGreen}{-0.020})$ & \cellcolor{green!4}$0.024(\textcolor{PineGreen}{-0.031})$ & \cellcolor{green!4}$0.020(\textcolor{PineGreen}{-0.012})$ \bigstrut[t]\\
  WV2  &\cellcolor{green!4}$0.081(\textcolor{PineGreen}{-0.022})$ & \cellcolor{green!4}$0.134(\textcolor{PineGreen}{-0.024})$ & \cellcolor{green!4}$0.137(\textcolor{PineGreen}{-0.049})$ & \cellcolor{green!4}$0.045(\textcolor{PineGreen}{-0.008})$ & \cellcolor{green!4}$0.071(\textcolor{PineGreen}{-0.012})$ & \cellcolor{green!4}$0.032(\textcolor{PineGreen}{-0.010})$ & \cellcolor{green!4}$0.085(\textcolor{PineGreen}{-0.018})$ & \cellcolor{green!4}$0.153(\textcolor{PineGreen}{-0.005})$ & \cellcolor{green!4}$0.133(\textcolor{PineGreen}{-0.053})$ & \cellcolor{green!4}$0.047(\textcolor{PineGreen}{-0.006})$ & \cellcolor{green!4}$0.071(\textcolor{PineGreen}{-0.012})$ & \cellcolor{green!4}$0.039(\textcolor{PineGreen}{-0.003})$ \bigstrut[t]\\
  WV3  &\cellcolor{green!4}$0.096(\textcolor{PineGreen}{-0.000})$ & \cellcolor{green!4}$0.141(\textcolor{PineGreen}{-0.027})$ & \cellcolor{green!4}$0.167(\textcolor{PineGreen}{-0.050})$ & \cellcolor{red!4}$0.054(\textcolor{BrickRed}{+0.002})$ & \cellcolor{green!4}$0.067(\textcolor{PineGreen}{-0.011})$ & \cellcolor{green!4}$0.033(\textcolor{PineGreen}{-0.004})$ & \cellcolor{green!4}$0.083(\textcolor{PineGreen}{-0.013})$ & \cellcolor{green!4}$0.134(\textcolor{PineGreen}{-0.033})$ & \cellcolor{green!4}$0.146(\textcolor{PineGreen}{-0.072})$ & \cellcolor{green!4}$0.043(\textcolor{PineGreen}{-0.008})$ & \cellcolor{green!4}$0.064(\textcolor{PineGreen}{-0.014})$ & \cellcolor{green!4}$0.033(\textcolor{PineGreen}{-0.003})$ \bigstrut[t]\\
  WV4  &\cellcolor{green!4}$0.053(\textcolor{PineGreen}{-0.009})$ & \cellcolor{green!4}$0.073(\textcolor{PineGreen}{-0.036})$ & \cellcolor{green!4}$0.099(\textcolor{PineGreen}{-0.029})$ & \cellcolor{green!4}$0.035(\textcolor{PineGreen}{-0.006})$ & \cellcolor{green!4}$0.053(\textcolor{PineGreen}{-0.011})$ & \cellcolor{green!4}$0.021(\textcolor{PineGreen}{-0.005})$ & \cellcolor{green!4}$0.045(\textcolor{PineGreen}{-0.018})$ & \cellcolor{green!4}$0.082(\textcolor{PineGreen}{-0.027})$ & \cellcolor{green!4}$0.076(\textcolor{PineGreen}{-0.052})$ & \cellcolor{green!4}$0.036(\textcolor{PineGreen}{-0.004})$ & \cellcolor{green!4}$0.052(\textcolor{PineGreen}{-0.012})$ & \cellcolor{green!4}$0.021(\textcolor{PineGreen}{-0.005})$ \bigstrut[t]\\
  \hline
  \multicolumn{1}{c}{\multirow{2}[4]{*}{$D_{\rho}\downarrow$}} & \multicolumn{6}{c}{$\rho_t=\mathcal{N}(0, 1)$} & \multicolumn{6}{c}{$\rho_t=\mathcal{U}(0, 1)$} \bigstrut[t]\\
  \cmidrule(r){2-7}  \cmidrule(r){8-13}      & \multicolumn{1}{c}{GF1} & \multicolumn{1}{c}{IK} & \multicolumn{1}{c}{QB} & \multicolumn{1}{c}{WV2} & \multicolumn{1}{c}{WV3} & \multicolumn{1}{c}{WV4} & \multicolumn{1}{c}{GF1} & \multicolumn{1}{c}{IK} & \multicolumn{1}{c}{QB} & \multicolumn{1}{c}{WV2} & \multicolumn{1}{c}{WV3} & \multicolumn{1}{c}{WV4} \bigstrut\\
  \hline
  GF1  &\cellcolor{red!4}$0.043(\textcolor{BrickRed}{+0.003})$ & \cellcolor{red!4}$0.110(\textcolor{BrickRed}{+0.030})$ & \cellcolor{green!4}$0.026(\textcolor{PineGreen}{-0.008})$ & \cellcolor{red!4}$0.079(\textcolor{BrickRed}{+0.032})$ & \cellcolor{red!4}$0.062(\textcolor{BrickRed}{+0.015})$ & \cellcolor{green!4}$0.041(\textcolor{PineGreen}{-0.002})$ & \cellcolor{green!4}$0.019(\textcolor{PineGreen}{-0.020})$ & \cellcolor{green!4}$0.059(\textcolor{PineGreen}{-0.021})$ & \cellcolor{green!4}$0.022(\textcolor{PineGreen}{-0.013})$ & \cellcolor{green!4}$0.015(\textcolor{PineGreen}{-0.032})$ & \cellcolor{green!4}$0.017(\textcolor{PineGreen}{-0.029})$ & \cellcolor{green!4}$0.017(\textcolor{PineGreen}{-0.026})$ \bigstrut[t]\\
  IK   &\cellcolor{green!4}$0.082(\textcolor{PineGreen}{-0.006})$ & \cellcolor{green!4}$0.095(\textcolor{PineGreen}{-0.038})$ & \cellcolor{green!4}$0.090(\textcolor{PineGreen}{-0.058})$ & \cellcolor{red!4}$0.126(\textcolor{BrickRed}{+0.024})$ & \cellcolor{green!4}$0.110(\textcolor{PineGreen}{-0.010})$ & \cellcolor{green!4}$0.061(\textcolor{PineGreen}{-0.008})$ & \cellcolor{green!4}$0.064(\textcolor{PineGreen}{-0.024})$ & \cellcolor{green!4}$0.092(\textcolor{PineGreen}{-0.040})$ & \cellcolor{green!4}$0.066(\textcolor{PineGreen}{-0.083})$ & \cellcolor{green!4}$0.051(\textcolor{PineGreen}{-0.049})$ & \cellcolor{green!4}$0.061(\textcolor{PineGreen}{-0.059})$ & \cellcolor{green!4}$0.045(\textcolor{PineGreen}{-0.024})$ \bigstrut[t]\\
  QB   &\cellcolor{green!4}$0.061(\textcolor{PineGreen}{-0.003})$ & \cellcolor{green!4}$0.054(\textcolor{PineGreen}{-0.020})$ & \cellcolor{green!4}$0.075(\textcolor{PineGreen}{-0.038})$ & \cellcolor{green!4}$0.052(\textcolor{PineGreen}{-0.026})$ & \cellcolor{green!4}$0.060(\textcolor{PineGreen}{-0.019})$ & \cellcolor{green!4}$0.044(\textcolor{PineGreen}{-0.017})$ & \cellcolor{green!4}$0.036(\textcolor{PineGreen}{-0.028})$ & \cellcolor{green!4}$0.048(\textcolor{PineGreen}{-0.026})$ & \cellcolor{green!4}$0.029(\textcolor{PineGreen}{-0.083})$ & \cellcolor{green!4}$0.021(\textcolor{PineGreen}{-0.057})$ & \cellcolor{green!4}$0.023(\textcolor{PineGreen}{-0.056})$ & \cellcolor{green!4}$0.028(\textcolor{PineGreen}{-0.034})$ \bigstrut[t]\\
  WV2  &\cellcolor{green!4}$0.166(\textcolor{PineGreen}{-0.027})$ & \cellcolor{green!4}$0.198(\textcolor{PineGreen}{-0.039})$ & \cellcolor{green!4}$0.199(\textcolor{PineGreen}{-0.059})$ & \cellcolor{red!4}$0.115(\textcolor{BrickRed}{+0.001})$ & \cellcolor{green!4}$0.130(\textcolor{PineGreen}{-0.045})$ & \cellcolor{green!4}$0.093(\textcolor{PineGreen}{-0.026})$ & \cellcolor{green!4}$0.164(\textcolor{PineGreen}{-0.029})$ & \cellcolor{green!4}$0.235(\textcolor{PineGreen}{-0.002})$ & \cellcolor{green!4}$0.200(\textcolor{PineGreen}{-0.058})$ & \cellcolor{green!4}$0.100(\textcolor{PineGreen}{-0.013})$ & \cellcolor{green!4}$0.126(\textcolor{PineGreen}{-0.048})$ & \cellcolor{green!4}$0.112(\textcolor{PineGreen}{-0.007})$ \bigstrut[t]\\
  WV3  &\cellcolor{red!4}$0.205(\textcolor{BrickRed}{+0.018})$ & \cellcolor{green!4}$0.214(\textcolor{PineGreen}{-0.041})$ & \cellcolor{green!4}$0.234(\textcolor{PineGreen}{-0.053})$ & \cellcolor{red!4}$0.231(\textcolor{BrickRed}{+0.109})$ & \cellcolor{red!4}$0.173(\textcolor{BrickRed}{+0.011})$ & \cellcolor{green!4}$0.105(\textcolor{PineGreen}{-0.015})$ & \cellcolor{green!4}$0.176(\textcolor{PineGreen}{-0.010})$ & \cellcolor{green!4}$0.226(\textcolor{PineGreen}{-0.029})$ & \cellcolor{green!4}$0.219(\textcolor{PineGreen}{-0.068})$ & \cellcolor{green!4}$0.117(\textcolor{PineGreen}{-0.004})$ & \cellcolor{green!4}$0.131(\textcolor{PineGreen}{-0.030})$ & \cellcolor{green!4}$0.101(\textcolor{PineGreen}{-0.018})$ \bigstrut[t]\\
  WV4  &\cellcolor{green!4}$0.105(\textcolor{PineGreen}{-0.001})$ & \cellcolor{green!4}$0.112(\textcolor{PineGreen}{-0.042})$ & \cellcolor{green!4}$0.115(\textcolor{PineGreen}{-0.044})$ & \cellcolor{red!4}$0.146(\textcolor{BrickRed}{+0.080})$ & \cellcolor{red!4}$0.169(\textcolor{BrickRed}{+0.077})$ & \cellcolor{green!4}$0.038(\textcolor{PineGreen}{-0.014})$ & \cellcolor{green!4}$0.072(\textcolor{PineGreen}{-0.034})$ & \cellcolor{green!4}$0.118(\textcolor{PineGreen}{-0.035})$ & \cellcolor{green!4}$0.105(\textcolor{PineGreen}{-0.055})$ & \cellcolor{green!4}$0.053(\textcolor{PineGreen}{-0.012})$ & \cellcolor{green!4}$0.062(\textcolor{PineGreen}{-0.029})$ & \cellcolor{green!4}$0.035(\textcolor{PineGreen}{-0.017})$ \bigstrut[t]\\
  \bottomrule
  \end{tabular}%
  }
  \caption{Average results of $12$ models. Rows correspond to training satellites, and columns correspond to testing satellites. \textcolor{ForestGreen}{Green} indicates performance improvement, while \textcolor{OrangeRed}{red} indicates performance degradation. The values in parentheses represent the average delta compared to the baseline without UniPAN.}
  \label{tab:cmp-all}%
  \vspace{-4mm}
  \end{table*}%

%% file: 04_experiment.tex
\section{Experiment}
\label{sec:exp}

\subsection{Setups}

\nbf{Datasets} We conducted experiments utilizing the Red, Green, Blue, and NIR bands of the NBUPansharpRSData dataset~\cite{surv2}. The variations in resolution among different satellites, coupled with significant differences in land covers and land uses, render this dataset particularly suitable for assessing the generalization capabilities of pansharpening models. The dataset comprises data from $6$ satellites. For each satellite, we partitioned the data into training, validation, and testing sets in a ratio of $7:1:2$.

\nbf{Implementation Details} To validate the effectiveness and generalization capability of UniPAN, we conducted comparative experiments between baseline (w/o UniPAN) and enhanced (w/ UniPAN) configurations using diverse commonly-recognized models, including FusionNet~\cite{FusionNet}, PanNet~\cite{PanNet}, MSDDN~\cite{MSDDN1,MSDDN2}, GPPNN~\cite{GPPNN}, MSDCNN~\cite{MSDCNN}, FeINFN~\cite{FeINFN}, PNN~\cite{PNN}, LAGConv~\cite{LAGConv}, SFIIN~\cite{SFIIN,SFIIN++}, PreMix~\cite{PreMix}, UAPN~\cite{UAPN}, MDCUN~\cite{MDCUN}. All implementations followed identical experimental procedures: a batch size of $8$ (except for MDCUN~\cite{MDCUN} and FeINFN~\cite{FeINFN} which used $4$), the Adam~\cite{Adam} optimizer ($\beta_1 = 0.9$, $\beta_2 = 0.999$) with an initial learning rate of $10^{-3}$, and StepLR scheduling which reduces the learning rate by a factor of $0.8$ every $100$ epochs. Training was performed on an NVIDIA GeForce RTX $4090$ GPU for $600$ epochs using the Wald protocol~\cite{wald} with $l_1$ loss, with best model selected based on optimal validation performance. Note that, to prevent data leakage of testing set, the transformation fitting process is conducted on the training set of each satellite.

\input{figs/cmp-1.tex}

\nbf{Evaluation Metrics} The full-reference metrics based on spatial degradation (\eg PSNR, SSIM) can only reflect performance in simulated reduced-resolution experiments and fail to demonstrate effectiveness in real-world full-resolution scenarios~\cite{9779258,Pan-GAN,PEMAE}. In contrast, no-reference metrics like QNR explicitly quantify model performance in both spectral preservation and spatial enhancement, enabling direct evaluation of fusion results at native resolution while avoiding theoretical biases introduced by downsampling. To better align with practical applications where reference images are unavailable, this paper adopts widely-used no-reference spectral and spatial metrics, \ie $\text{QNR}$~\cite{QNR}, ${D}_{\lambda}$~\cite{QNR}, ${D}_S$~\cite{QNR}, and ${D}_{\rho}$~\cite{drou}, to comprehensively assess cross-resolution and cross-satellite generalization capabilities.

\subsection{Quantitative Evaluation}

\cref{tab:cmp-all} presents the average generalization performance of various models enhanced with UniPAN under different target distributions. \textcolor{PineGreen}{Green} indicate metric improvements, while \textcolor{BrickRed}{red} denote performance degradation. Slight yet observable differences exist between the two target distributions. Overall, the uniform distribution achieves better performance than the normal distribution. The normal distribution demonstrated a modest decline in performance across the WV2 and WV3 datasets. This phenomenon may stem from the peaked characteristics of the normal distribution, which could hinder model training, whereas the uniform distribution provides smoother normalization (see \cref{fig:kde} in \cref{sec:visual}).

Notably, UniPAN with $\rho_t = \mathcal{U}(0,1)$ achieves significant improvements across most metrics, with only marginal declines in rare cases. For instance, when generalizing from WV3 to QB, UniPAN improves the QNR metric by over $0.084$ (from $0.729$ to $0.813$) averaged on all $12$ models, underscoring its effectiveness. Beyond QNR, UniPAN also enhances spectral preservation ($D_\lambda$) and spatial consistency ($D_S, D_\rho$), further validating the efficacy of the unified distribution strategy. These results confirm that aligning heterogeneous satellite data to a unified distribution effectively mitigates domain shifts, enabling robust cross-sensor generalization. However, while UniPAN improves spatial fidelity greatly, it may cause slight spectral distortion, as evidenced by the minor performance degradation observed in the $D_{\lambda}$ metric. This is reasonable since the transformation applied to each spectral band maintains the relative spatial structure but might result in minimal loss of spectral information. Nevertheless, such information loss is extremely limited, as the overall enhancement in performance remains predominant.

\input{figs/kde-qq-before-after.tex}

\subsection{Visual Comparison}
\label{sec:visual}

\nbf{Comparison of Fusion Results} To illustrate the practical impact of UniPAN, \cref{fig:vis-cmp1,fig:vis-cmp2} compare fusion results w/ and w/o UniPAN. In most cases, UniPAN achieves highly favorable results in both spectral and spatial preservation. Furthermore, the remarkably high QNR values demonstrate its superior capability in balancing spectral and spatial fidelity, \eg in~\cref{fig:vis-cmp1}, MDCUN~\cite{MDCUN} applying UniPAN with normal distribution enhanced the QNR by $0.1514$ (from $0.7157$ to $0.8671$) and aligned fused results closely with the ground truth spectra. In the full-resolution generalization test depicted in~\cref{fig:vis-cmp2}, even with minimal visual differences, UniPAN achieves metrically significant improvements in spectral and spatial preservation. This capability is critical for downstream applications requiring accurate spectral information. Additional visual comparison results are included in the supplementary materials. 

\input{figs/abl-nq.tex}

\nbf{Comparison of Data Distribution} \cref{fig:kde} further analyzes data distribution alignment. The kernel density estimation (KDE) curves (top right area and diagonals) reveal that UniPAN successfully transforms the original complex and heterogeneous distributions into a unified target distribution. The quantile-quantile (Q-Q) plots (bottom left area) demonstrate harmonized distributions across satellites after applying UniPAN (The greater the deviation from the $y=x$ line, the more significant the disparity between the two distributions). For example, the original Q-Q plots between WV4 and QB show significant dispersion, whereas post-UniPAN points close tightly around $y=x$. This visual evidence corroborates UniPAN's ability to bridge distribution gaps.

\input{figs/abl-nsub.tex}

\subsection{Parameter Analysis}

UniPAN introduces two hyperparameters: the number of sampled pixels $n$ and the number of quantiles $m$. \cref{fig:param1,fig:param2} investigate their impact on performance of FusionNet~\cite{FusionNet} trained on WV2 and tested on GF1. As $n$ and $m$ vary, the metrics stabilize. Generally, $\mathcal{U}(0,1)$ demonstrates superior performance compared to $\mathcal{N}(0,1)$. Both configurations achieve statistically significant improvements over baseline methods without applying UniPAN. This parametric robustness demonstrates that optimal performance can be attained without exhaustive hyperparameter tuning, significantly reducing computational overhead.

%% file: figs/cmp-1.tex
\begin{figure*}[t]
    \centering
    \vspace{-2mm}
    \includegraphics[width=0.96\linewidth]{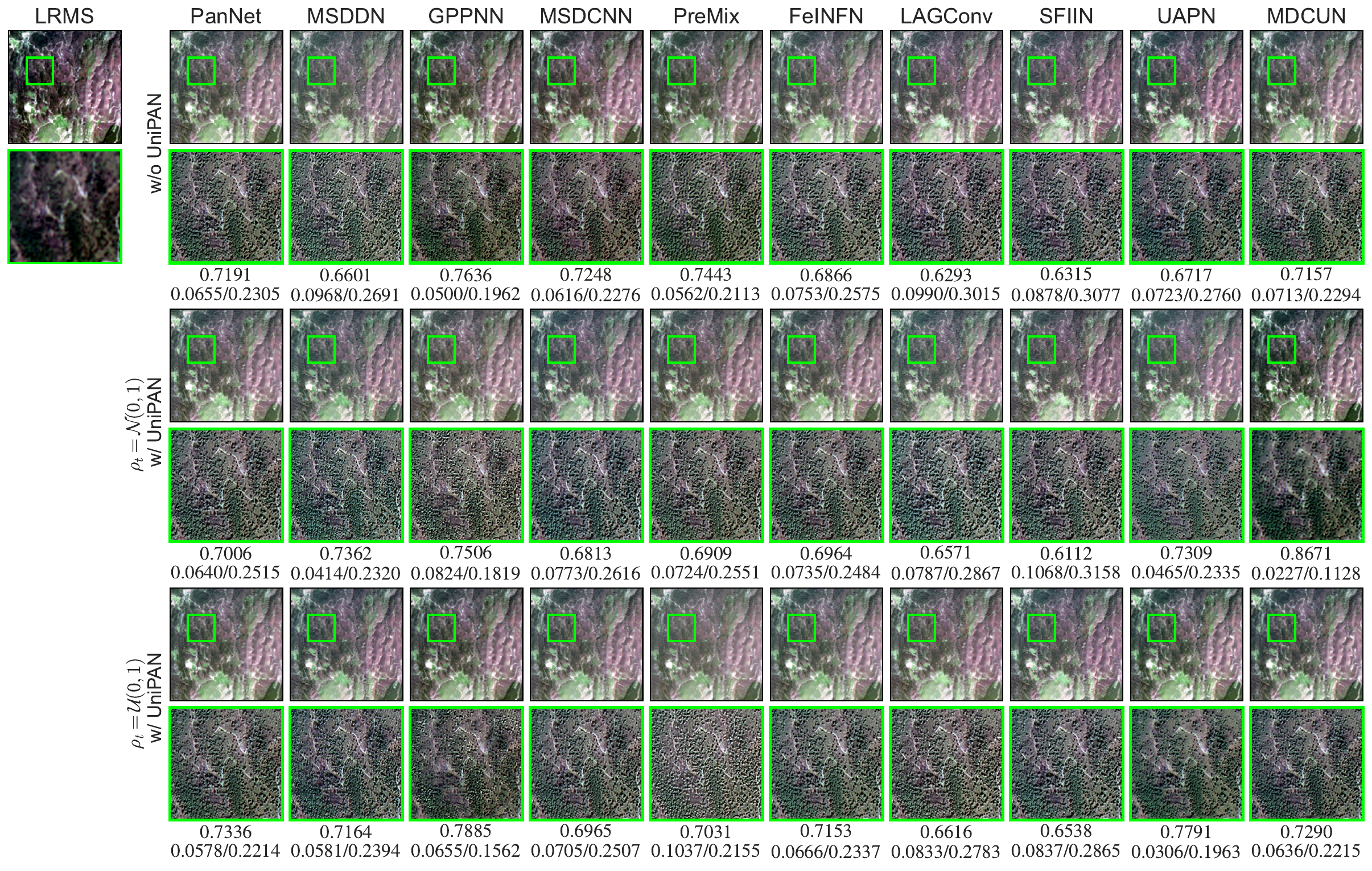}
    % \vspace{-5mm}
    % \caption{Generalization testing on Satellite B, the networks are trained on Satellite A. The best QNR value is highlighted in \textcolor{OrangeRed}{red}.}
    % \vspace{-1mm}
    \caption{Generalization testing on GaoFen-1, the networks are trained on WorldView-3. The \textcolor{green}{bright green} box indicates the zoomed-in area. Below the zoomed-in image, the first line is the QNR value, and the second line is the $D_{\lambda}$/$D_S$ value. In most cases, the use of UniPAN yields superior spectral and spatial preservation. For instance, MDCUN~\cite{MDCUN} exhibits a highly close alignment with the LRMS in spectral characteristics with UniPAN of $\rho_t = \mathcal{N}(0,1)$.}
    \label{fig:vis-cmp1}
    \vspace{-3mm}
\end{figure*}

% In most cases, the use of UniPAN yields superior spectral and spatial preservation. For instance, MDCUN~\cite{MDCUN} exhibits a highly close alignment with the LRMS in spectral characteristics with UniPAN of $\rho_t = \mathcal{N}(0,1)$.

% Use figure* for multi-column figure
\begin{figure*}[t]
    \centering
    \vspace{-2mm}
    \includegraphics[width=0.96\linewidth]{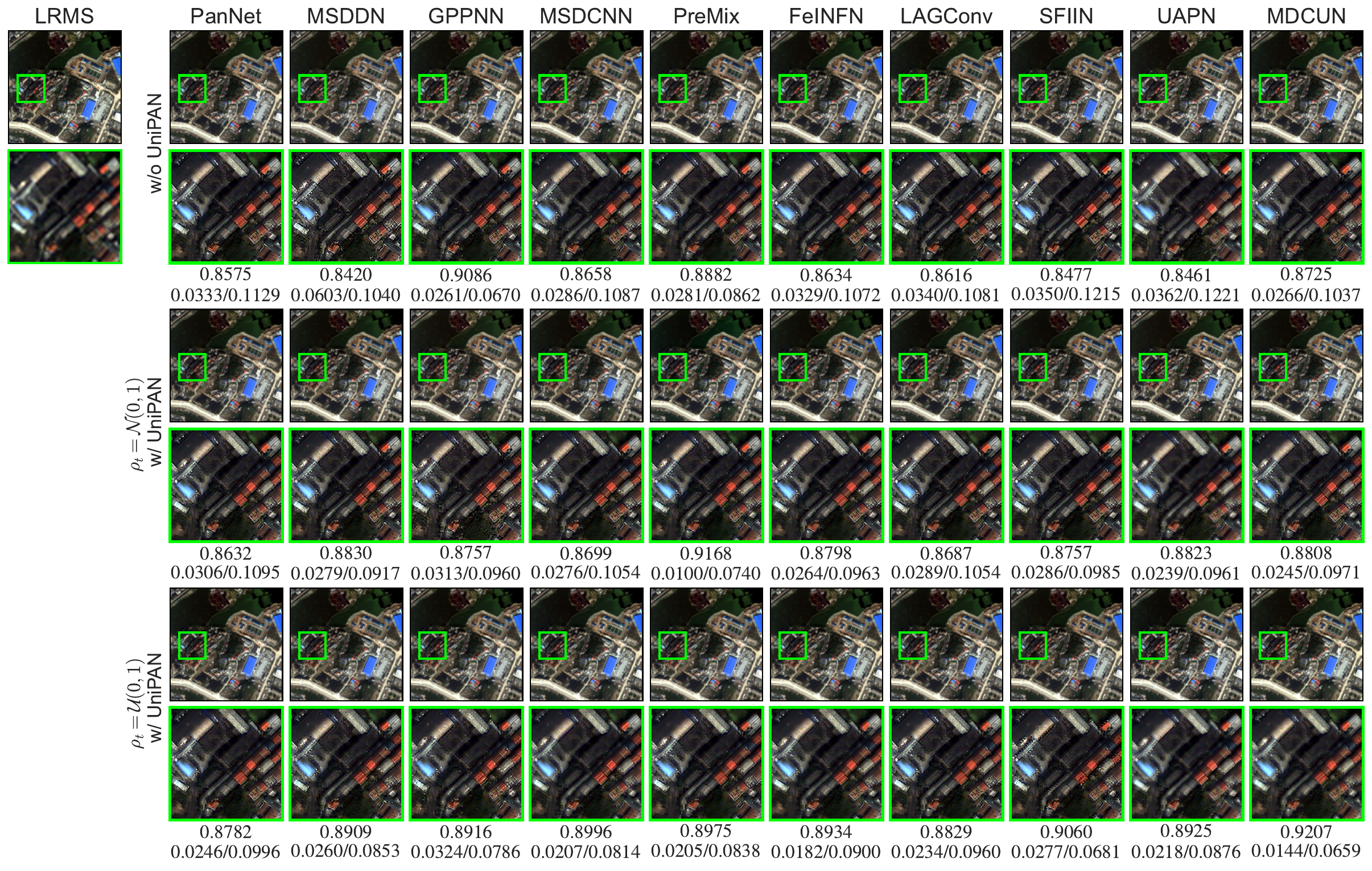}
    % \vspace{-5mm}
    % \caption{Generalization testing on Satellite B, the networks are trained on Satellite A. The best QNR value is highlighted in \textcolor{OrangeRed}{red}.}
    \caption{Full-resolution generalization testing on QuickBird. The \textcolor{green}{bright green} box indicates the zoomed-in area. Below the zoomed-in image, the first line is the QNR value, and the second line is the $D_{\lambda}$/$D_S$ value.}
    \label{fig:vis-cmp2}
    \vspace{-3mm}
\end{figure*}

%% file: figs/kde-qq-before-after.tex
\begin{figure*}[t]
    \centering
    % \vspace{-2mm}
    \begin{subfigure}[b]{0.32\linewidth}
        \includegraphics[width=\linewidth]{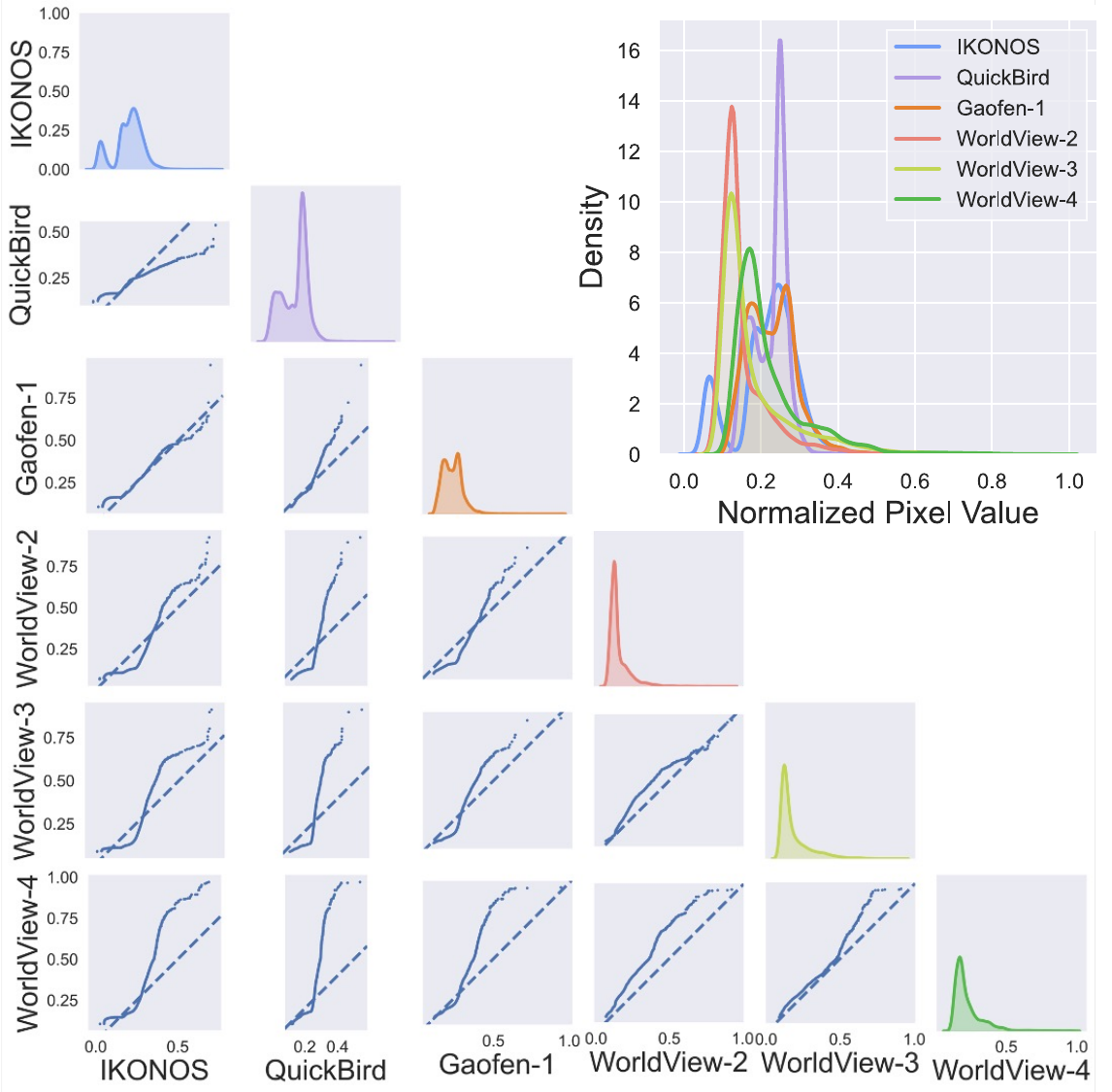}
        \caption{w/o UniPAN}
        \label{fig:subfigA}
    \end{subfigure}
    \hfill
    \begin{subfigure}[b]{0.32\linewidth}
        \includegraphics[width=\linewidth]{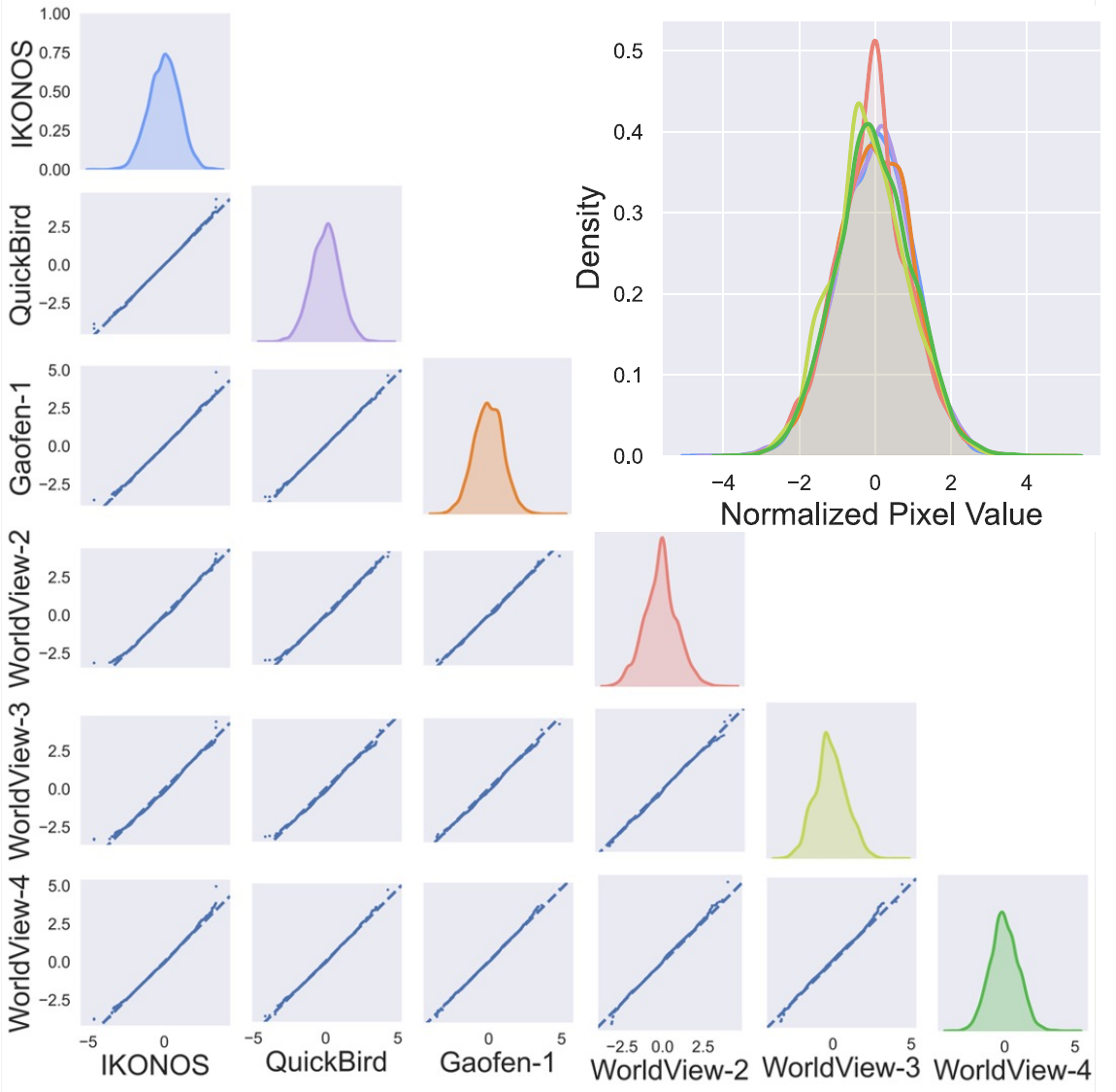}
        \caption{w/ UniPAN, $\rho_t=\mathcal{N}(0,1)$}
        \label{fig:subfigB}
    \end{subfigure}
    \hfill
    \begin{subfigure}[b]{0.32\linewidth}
        \vspace{1mm}
        \includegraphics[width=\linewidth]{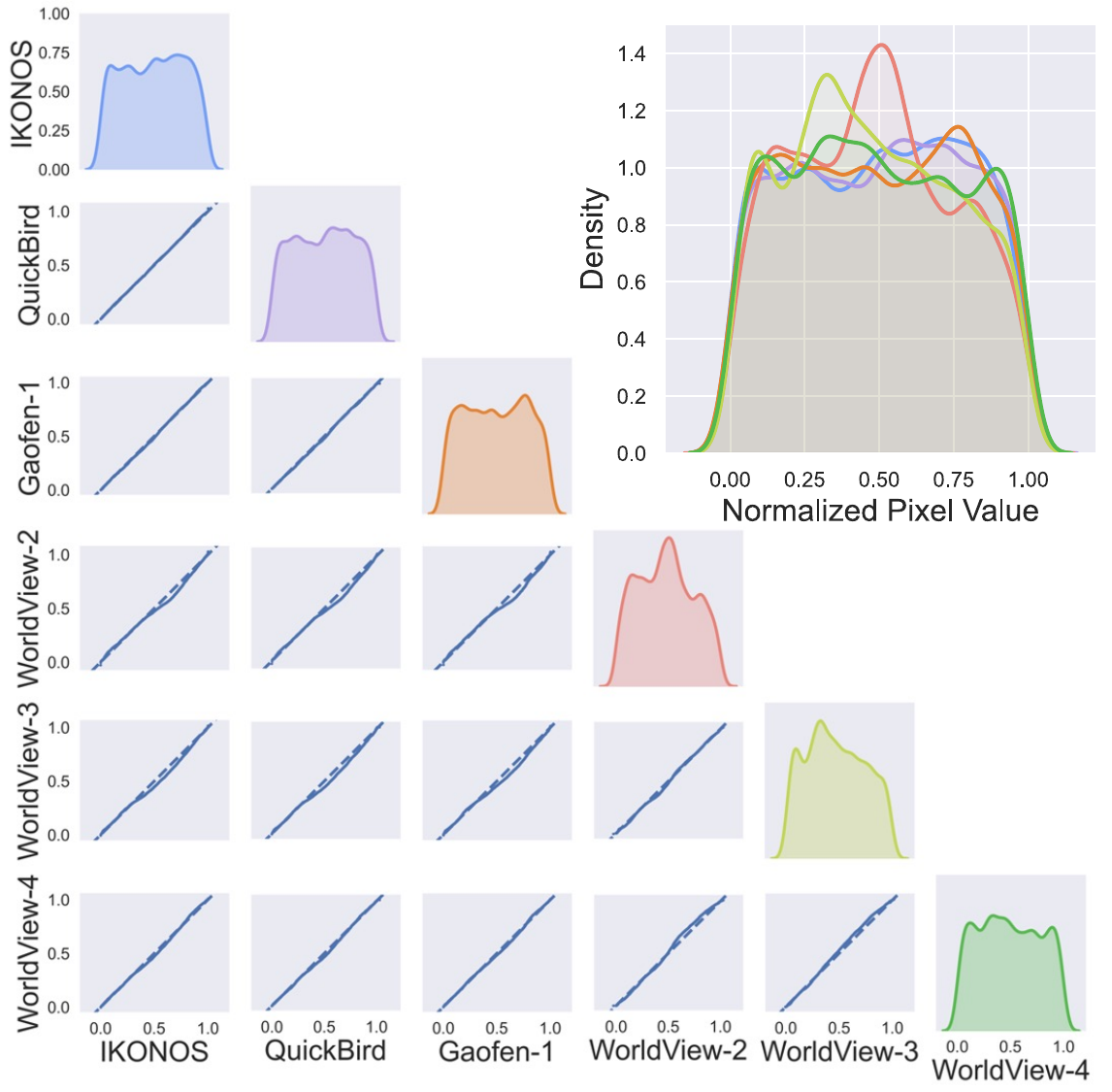}
        \caption{w/ UniPAN, $\rho_t=\mathcal{U}(0,1)$}
        \label{fig:subfigC}
    \end{subfigure}
    \vspace{-2mm}
    \caption{Comparison of data distributions before and after applying UniPAN. The upper-right and diagonals of each subplot display the kernel density estimation (KDE) curves of Green band for each satellite. The scatter plots in the lower-left areas of each subplots represent quantile-quantile (Q-Q) plots of green band, with the dashed line indicating $y=x$. The greater the deviation from the $y=x$ line, the more significant the disparity between the two distributions. Zoom in for better view.}
    \label{fig:kde}
    \vspace{-2mm}
\end{figure*}

%% file: figs/abl-nq.tex
% Use figure* for multi-column figure
\begin{figure}[t]
    \centering
    \vspace{-3mm}
    \hspace{-5mm}
    \includegraphics[width=\linewidth]{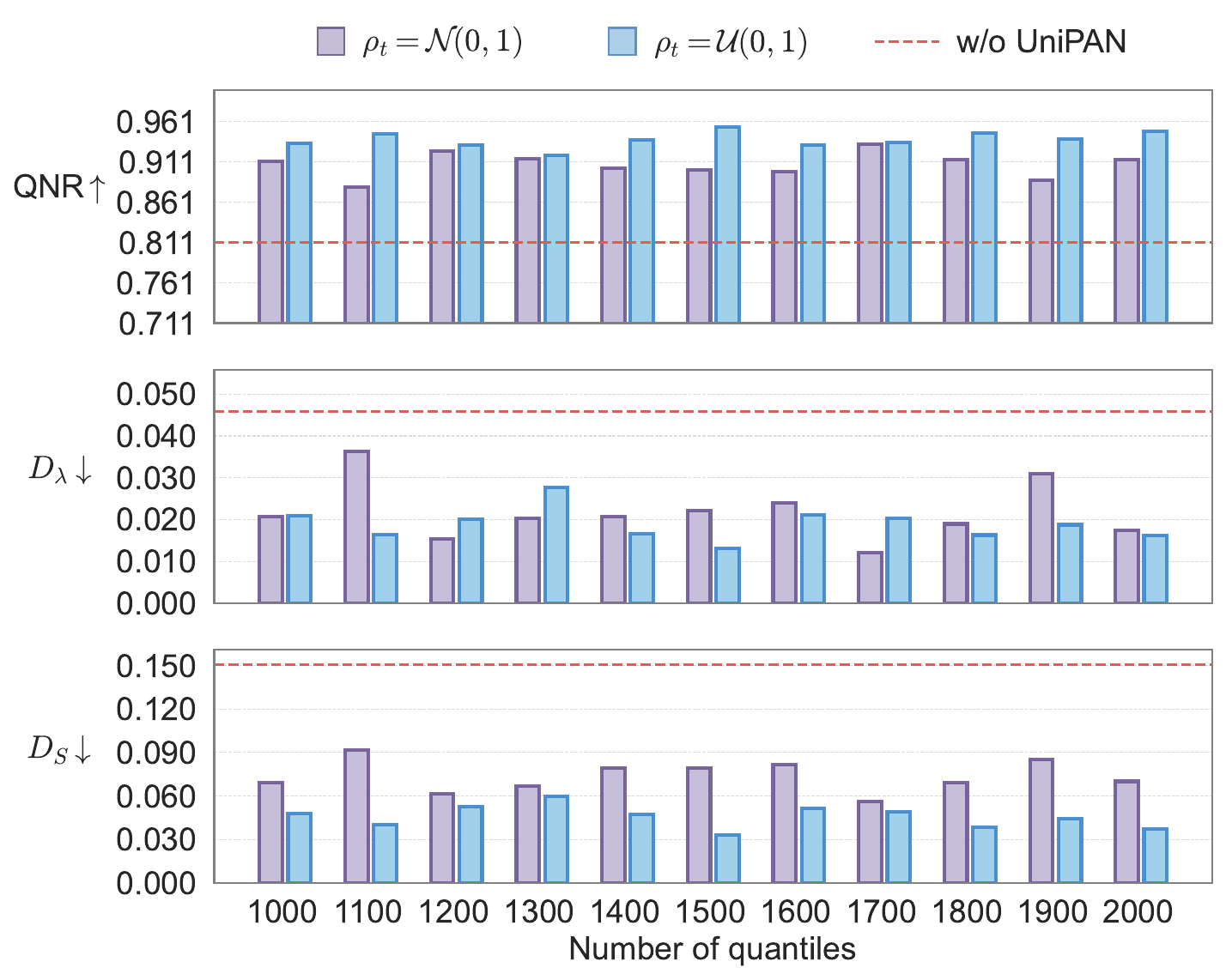}
    \vspace{-2mm}
    \caption{Parameter analysis of number of quantiles, with number of sampled pixels fixed to $10000$.}
    \label{fig:param1}
    \vspace{-4mm}
\end{figure}

%% file: figs/abl-nsub.tex
% Use figure* for multi-column figure
\begin{figure}[t]
    \centering
    \vspace{-3mm}
    \hspace{-5mm}
    \includegraphics[width=\linewidth]{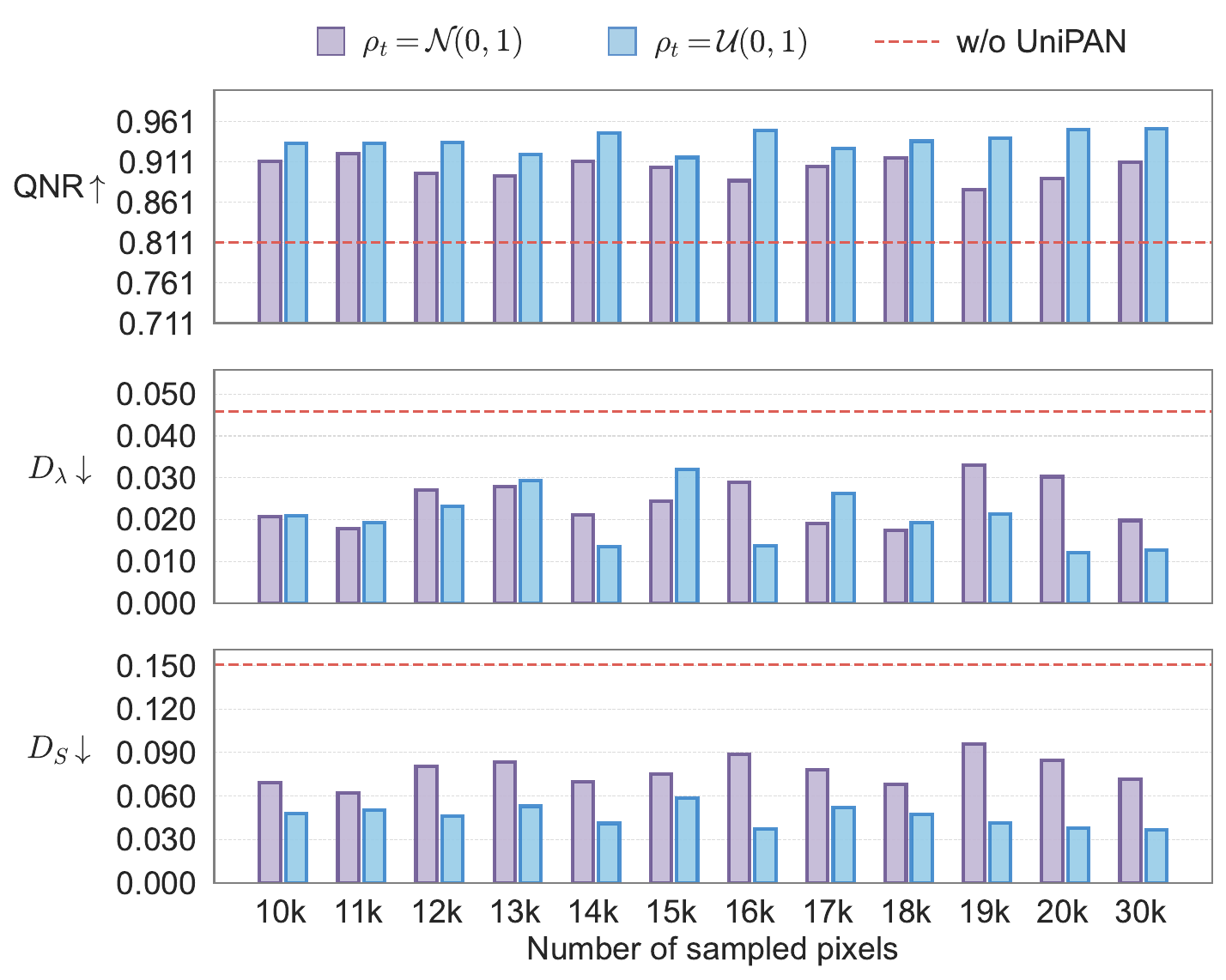}
    \vspace{-2mm}
    \caption{Parameter analysis of number of sampled pixels , with number of quantiles fixed to $1000$.}
    \label{fig:param2}
    \vspace{-4mm}
\end{figure}

%% file: 10_conclusion.tex
\section{Conclusion and Discussion}
\label{sec:conclusion}

We proposed UniPAN, a unified distribution strategy addressing generalization challenges in deep learning-based pansharpening. UniPAN eliminates spectral discrepancies by normalizing multi-source remote sensing imagery into predefined target distributions, effectively bridging the gap between training and testing distributions. Extensive quantitative and visual comparisons across $6$ satellite sensors and $12$ models demonstrate that UniPAN significantly enhances cross-sensor generalization. Notably, this enhancement is achieved without introducing any additional trainable parameters or the need for retraining or fine-tuning the models.

While this work aligns theoretically with 1D optimal transport, considering interdependencies of spectral bands or spatial patches, future research could explore extending the method to the multi-dimensional case with optimal transport theories. Additionally, the experimental results indicate that the choice of target distribution has a notable impact on performance, thus it is worthwhile investigating adaptive target domain selection. To further validate the generalizability of the proposed unified distribution strategy, the ideas presented in this work could be extended to other remote sensing image processing tasks, such as multi-modal fusion and hyperspectral pansharpening, \etc